\documentclass[10pt,journal,compsoc]{IEEEtran}

\ifCLASSOPTIONcompsoc
\usepackage[nocompress]{cite}
\usepackage{lineno,hyperref}
\usepackage{algorithmic}
\usepackage{algorithm}

\usepackage{graphicx}
\usepackage{booktabs}
\usepackage{stfloats}
\usepackage{threeparttable}
\usepackage{tabu}
\usepackage{url}

\usepackage{epsfig}
\usepackage{amsmath,amssymb,amsfonts,latexsym}
\usepackage{colortbl,multirow,hhline}

\usepackage{algorithm}
\usepackage{array,color}
\usepackage{CJK,graphicx}

\usepackage{subfigure}
\usepackage{slashbox}
\usepackage{colortbl,multirow,hhline}
\usepackage{color}
\usepackage[table,xcdraw]{xcolor}
\usepackage{lscape}
\usepackage{epstopdf}
\usepackage[normalem]{ulem} 

\else
  \usepackage{cite}
\fi
\ifCLASSINFOpdf
\else
\fi
\begin{document}
%
\title{CAGFuzz: Coverage-Guided Adversarial Generative Fuzzing Testing of\\ Deep Learning Systems
}
\author{Pengcheng Zhang, \emph{Member, IEEE},
        Qiyin Dai,
        Patrizio Pelliccione
\IEEEcompsocitemizethanks{\IEEEcompsocthanksitem P. Zhang and Q. Dai are with the College of Computer and Information,
Hohai University, Nanjing, P.R.China\protect\\
E-mail: pchzhang@hhu.edu.cn
\IEEEcompsocthanksitem P. Pelliccione is with the University of L'Aquila, Italy and Chalmers $|$ University of Gothenburg, Sweden.\protect\\
E-mail: patrizio.pelliccione@univaq.it

}
\thanks{Manuscript received XXXX XXXX; revised XXXX, XXXX.}}

%
%

 \markboth{IEEE TRANSACTIONS ON Software Engineering, ~Vol.~XX, No.~X, XXXX}
 {Shell \MakeLowerexample{\textit{et al.}}: Bare Demo of IEEEtran.cls for Computer Society Journals}
%



\IEEEtitleabstractindextext{
\begin{abstract}
Deep Learning systems (DL) based on Deep Neural Networks (DNNs) are increasingly being used in various aspects of our life, including unmanned vehicles, speech processing, intelligent robotics and etc. Due to the limited dataset and the dependence on manually labeled data, DNNs always fail to detect erroneous behaviors. This may lead to serious problems. Several approaches have been proposed to enhance adversarial examples for testing DL systems. However, they have the following two limitations.
First, most of them do not consider the influence of small perturbations on adversarial examples. Some approaches take into account the perturbations, however, they design and generate adversarial examples based on special DNN models. This might hamper the reusability on the examples in other DNN models, thus reducing their generalizability.
Second, they only use shallow feature constraints (e.g. pixel-level constraints) to judge the difference between the generated adversarial example and the original example. The deep feature constraints, which contain high-level semantic information - such as image object category and scene semantics, are completely neglected.
To address these two problems, we propose \emph{CAGFuzz}, a \underline{C}overage-guided \underline{A}dversarial \underline{G}enerative \underline{Fuzz}ing testing approach for Deep Learning Systems, which generates adversarial examples for DNN models to discover their potential defects. First, we train an Adversarial Case Generator (\emph{AEG}) based on general data sets. \emph{AEG} only considers the data characteristics, and avoids low generalization ability.
Second, we extract the deep features of the original and adversarial examples, and constrain the adversarial examples by cosine similarity to ensure that the semantic information of adversarial examples remains unchanged. Finally, we use the adversarial examples to retrain the model.
Based on three standard data sets, we design a set of dedicated experiments to evaluate \emph{CAGFuzz}. The experimental results show that \emph{CAGFuzz} can improve the neuron coverage rate, detect hidden errors, and also improve the accuracy of the target DNN.

\end{abstract}

}

\maketitle

\IEEEdisplaynontitleabstractindextext

%
\IEEEpeerreviewmaketitle

\section{Introduction}\label{sec:introduction}
\IEEEPARstart{N}owadays, we have already stepped into the era of artificial intelligence from the digital era. Apps with AI systems can be seen everywhere in our daily life, such as Amazon Alexa~\cite{Lei2018The}, DeepMind's Atari~\cite{Volodymyr2015Human}, and AI-phaGo~\cite{Silver2016Mastering}.
 With the development of edge computing, 5G technology and etc., AI technologies become more and more mature. In many applications, we can see the shape of deep neural networks (DNNs), such as automatic driving~\cite{Sermanet2011Traffic}, intelligent robotics~\cite{Zhang2015Towards}, smart city applications~\cite{Wu2016Google} and AI-enabled Enterprise Information Systems~\cite{latah2018artificial}.
In this paper, we term this kind of applications as DL (deep learning) systems.

In particular, many different kinds of DNNs are embedded in security and safety-critical applications, such as automatic driving~\cite{Sermanet2011Traffic} and intelligent robotics~\cite{Zhang2015Towards}. This brings new challenges since predictability, correctness, and safety are especially crucial for this kind of DL systems.
These safety-critical applications deploying DNNs without comprehensive testing could have serious problems. For example, in automatic driving systems, if the deployed DNNs have not recognized the obstacles ahead timely and correctly, it may lead to serious consequences such as vehicle damage and even human death~\cite{Eykholt2018Robust}.

\begin{figure}[t]
 \centering
 \includegraphics[width=0.8\linewidth]{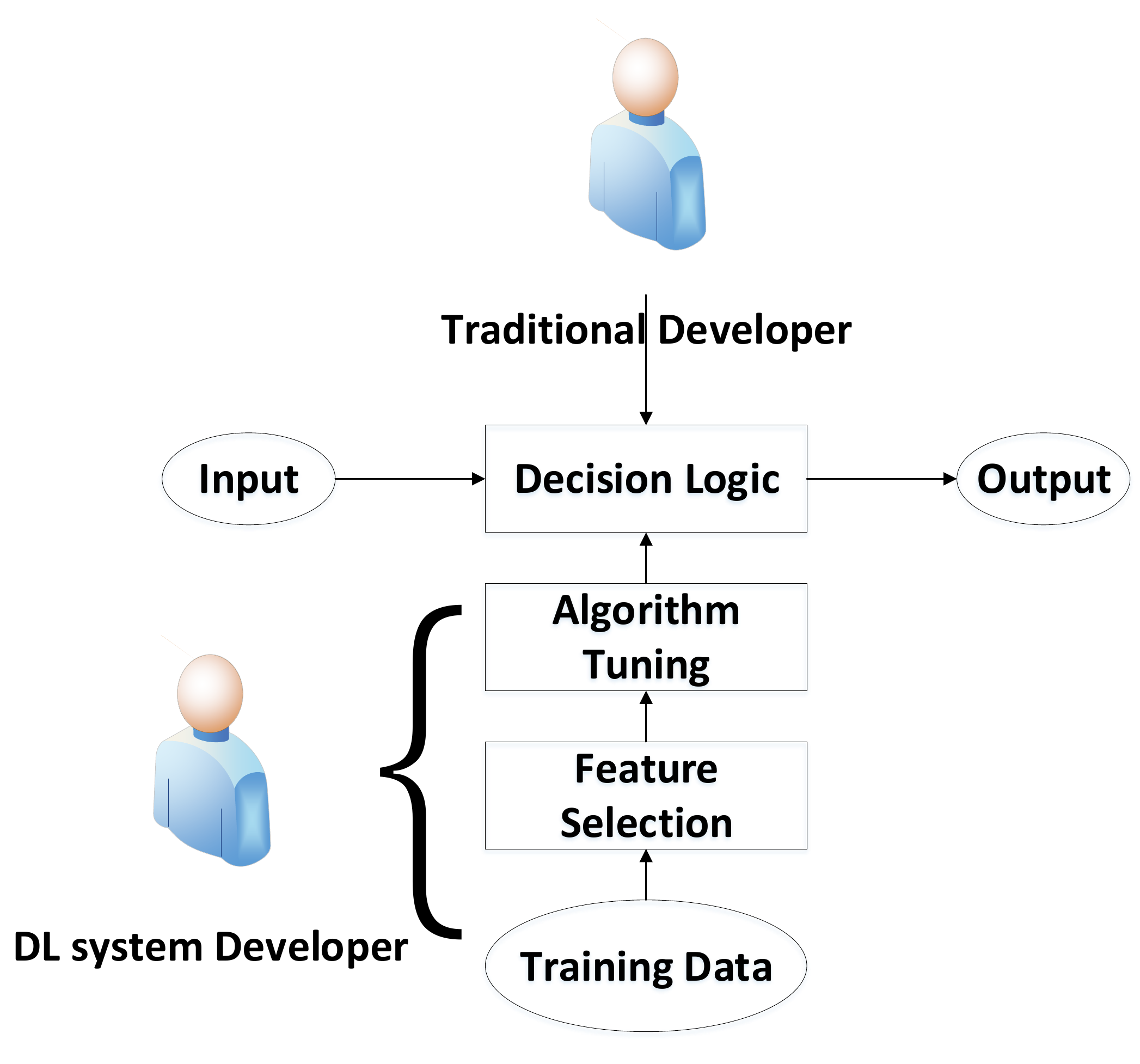}\\
 \caption{Comparison between traditional and DL system development}\label{Fig1}
\end{figure}

Generally speaking, the development process of DL systems is essentially different from the traditional software development process. As shown in Fig.~\ref{Fig1}, for traditional software development practices, developers directly specify the logic of the systems. On the contrary, DL systems automatically learn their models and corresponding parameters from data.
Consequently, the testing process of DL systems is also different from traditional software systems.
For traditional software systems, code or control-flow coverage is utilized to guide the testing process~\cite{bertolino2007software}. However, the logic of the DL systems is not encoded by control flow, and it cannot be solved by the normal encoding way. Their decisions are always made by training data for many times, and the performance is more dependent on data rather than human intervention. For DL systems, neural coverage can be used to guide the testing process~\cite{Pei2017DeepXplore}. When faults are found, it is also very difficult to locate the exact position in the original DL systems.
Consequently, most traditional software testing methodologies are not suitable for testing DL systems. As highlighted in~\cite{Pei2017DeepXplore,Xie2018DeepHunter}, research on developing new testing techniques for DL systems is urgently needed.

The standard way to test DL systems is to collect and manually mark as much actual test data as possible~\cite{fei2010imagenet,merkel2018software}. Obviously, it is unthinkable to exhaustively test every feasible input of the DL systems. Recently, an increasing number of researchers have contributed to test DL systems with a variety of approaches~\cite{Pei2017DeepXplore, Xie2018DeepHunter,guo2018dlfuzz, Sun2018Testing, Lei2018DeepGauge}. The main idea of these approaches is to enhance input examples of test data set by different techniques.
Some approaches, e.g. \emph{DeepXplore}~\cite{Pei2017DeepXplore}, use multiple DNNs to discover and generate adversarial examples that lie between the decision boundaries of these DNNs.
Some approaches, e.g. \emph{DeepHunter}~\cite{Xie2018DeepHunter}, use metamorphic mutation strategy to generate new test examples. Other approaches, e.g. \emph{DeepGauge}~\cite{Lei2018DeepGauge}, propose new coverage criteria for deep neural networks. These coverage criteria can be used as guidance for generating test examples. While state-of-the-art approaches make some progresses on testing DL systems, they still suffer the following two main problems:
\begin{enumerate}
  \item  
  \emph{DNN-dependent generation of adversarial examples.}
  Most approaches~\cite{Xie2018DeepHunter,tian2018deeptest} do not consider the influence of small perturbations on deep neural networks when the test examples are generated. Some approaches~\cite{Pei2017DeepXplore,goodfellow2014explaining} consider small perturbations based on special DNN models. The test examples that they have generated are only designed for one special DNN, and it may be difficult to generalize them to 
  other DNNs. Recent research on adversarial DL systems~\cite{Nguyen2015Deep,cubuk2017intriguing} shows that by adding the small perturbations to existing images and elaborating synthetic images can fool state-of-the-art DL systems. Therefore, to improve the generalization ability, it is significantly important to add small perturbations only based on data.
  \item \emph{Shallow feature constraints.} State-of-the-art adversarial example generation approaches use shallow feature constraints, such as pixel-level constraints, to judge the difference between the adversarial example and the original example. The deep feature constraints containing high-level semantic information, such as image object category and scene semantics, are completely neglected. For example, in their study, Xie et al.~\cite{Xie2018DeepHunter} use $L_{0}$ and $L_{\infty }$ to limit the pixel-level changes of the adversarial example. However, such shallow feature constraints can only represent the visual consistency between the adversarial example and the original example, and cannot guarantee the high-level semantic information consistency between the adversarial example and the original example. Furthermore, this may lead to bad performance when testing the network with deep layers.
\end{enumerate}


To address the problems aforementioned, we propose a new testing approach for DL systems, called \emph{CAGFuzz} (\underline{C}overage-guided \underline{A}dversarial \underline{G}enerative \underline{Fuzz}ing)\footnote{https://github.com/QXL4515/CAGFuzz}. The goal of the \emph{CAGFuzz} is to maximize the neuron coverage and generate adversarial test examples as much as possible with small perturbations for the target DNNs. The goal of the \emph{CAGFuzz} is to maximize the neuron coverage and generate adversarial test examples with minimal perturbations for the target DNNs. Meanwhile, the generated examples have strong generalization ability and can be used to test different DNN models. 
\emph{CAGFuzz} iteratively selects the test examples in the processing pool and generates the adversarial examples through the pre-trained adversarial example generator (see Section~\ref{sec_approach} for details) to guide DL systems to expose incorrect behaviors.
During the process of generating adversarial examples, \emph{CAGFuzz} keeps valid adversarial examples, which can provide a certain improvement in neuron coverage for subsequent fuzzy processing, and limit the small perturbations invisible to human eyes, ensuring the same meaningfulness between the original example and the adversarial example. The contributions of this paper include the following three aspects:


\begin{itemize}
   \item \emph{We design an adversarial example generator, AEG, which can generate adversarial examples with small perturbations based on general data sets.} 
   The goal of \emph{CycleGAN}~\cite{Zhu2017Unpaired} is to transform \emph{image A} to \emph{image B} with different styles. Based on \emph{CycleGAN}, our goal is to transform \emph{image B} back to \emph{image A}, and get \emph{image A'} similar to the original \emph{image A}. Consequently, we combine two generators with opposite functions of \emph{CycleGAN} as our adversarial example generator. The adversarial examples generated by \emph{AEG} can add small perturbations invisible to human eyes to the original examples.
   \emph{AEG} is trained based on general data sets and does not rely on any specific DNN model, which has higher generalization ability than state-of-the-art approaches. Furthermore, because of the inherent constraint logic of \emph{CycleGAN}, the trained \emph{AEG} not only has high efficiency in generating adversarial examples but also can effectively improve the robustness of DL systems.
   \item  \emph{We extract the deep features of the original example and the adversarial example, and make them as similar as possible by similarity measurement.} We use VGG-19 network~\cite{Simonyan2014Very} to extract the deep semantic information of the original example and the adversarial example, and use the method of cosine similarity measurement to ensure that the deep semantic information of the adversarial example is consistent with the original example as much as possible.
   At the same time, the deep feature constraint can make the adversarial examples generated by \emph{CAGFuzz} get better results compared with other approaches when testing the network with deep layers.
   \item \emph{We design a series of experiments to evaluate the CAGFuzz approach based on several public data sets.} The experiments validate that \emph{CAGFuzz} can effectively improve the neuron coverage of the target DNN model. Meanwhile, it is proved that the adversarial examples generated by \emph{CAGFuzz} can find hidden defects in the target DNN model. Furthermore, the accuracy and the robustness of the DNN models retrained by \emph{AEG} have been significantly improved. For example, the accuracy of the VGG-16~\cite{Simonyan2014Very} model in the experiments has been improved from the original 86.72\% to 97.25\%, with an improvement of 12.14\%. 
 \end{itemize}

The rest of the paper is organized as follows.
Section~\ref{sec_Preliminaries} provides some basic concepts including \emph{CycleGAN}, Coverage-guided Grey-box Fuzzing (CGF). The coverage-guided adversarial generative fuzzy testing framework is provided in Section~\ref{sec_approach}.
In Section~\ref{sec_validation}, we use three popular datasets (MNIST~\cite{Deng2012The}, Cifar-10~\cite{Li2017CIFAR10}, and ImageNet~\cite{Russakovsky2015ImageNet}) to validate our approach. Existing work and their limitations are discussed in Section~\ref{sec_Related work}. Finally, Section~\ref{sec_conclusions} concludes the paper and looks into future work.


\section{Preliminaries}\label{sec_Preliminaries}
The principles of coverage-guided grey-box fuzzing, \emph{CycleGAN}, and VGG-19 are introduced in Section~\ref{subsec_covfuzz}, Section~\ref{subsec_AEG}, and Section~\ref{subsec_featureEx}, respectively. Section~\ref{subsec_NC} introduces the basic concept and calculation formula for neuron coverage.
\subsection{Coverage-guided Grey-box Fuzzing}\label{subsec_covfuzz}
Due to the scalability and effectiveness in generating useful defect detection tests, fuzzing has been widely used in academia and industry. Based on the perception of the target program structure, the fuzzy controller can be divided into black-box, white-box, and grey-box. One of the most successful techniques is Coverage-guided Grey-box Fuzzing (CGF), which balances effectiveness and efficiency by using code coverage as feedback. Many state-of-the-art CGF approaches, such as AFL~\cite{zalewski2017american}, libFuzzer~\cite{serebryany2015libfuzzer}, and VUzzer~\cite{rawat2017vuzzer}, have been widely used and proved to be effective. Smart Greybox Fuzzing (SGF)~\cite{SGF} has made some improvements on CGF. Specifically, it leverages a high-level structural representation of the original example to generate new examples. The state-of-the-art CGF approaches mainly consist of three parts: \emph{mutation}, \emph{feedback guidance}, and \emph{fuzzing strategy}:
\begin{itemize}
\item \emph{Mutation:} According to the difference of the target application program and data format, the corresponding test data generation method is chosen and it can use the pre-generated examples, a variation of valid data examples, or dynamically generated ones according to the protocol or file format.
\item \emph{Feedback guidance:} The fuzzy test example is executed, and the target program is executed and monitored. The test data that causes the exception of the target program is recorded.
\item \emph{Fuzzing strategy:} If an error is detected, the corresponding example is reported and new generated examples that cover new traces are stored in the example pool.
\end{itemize}

\subsection{CycleGAN} \label{subsec_AEG}

\begin{figure}
\centering
   \subfigure[]{
   \label{figurea}
   \includegraphics[width=1\linewidth]{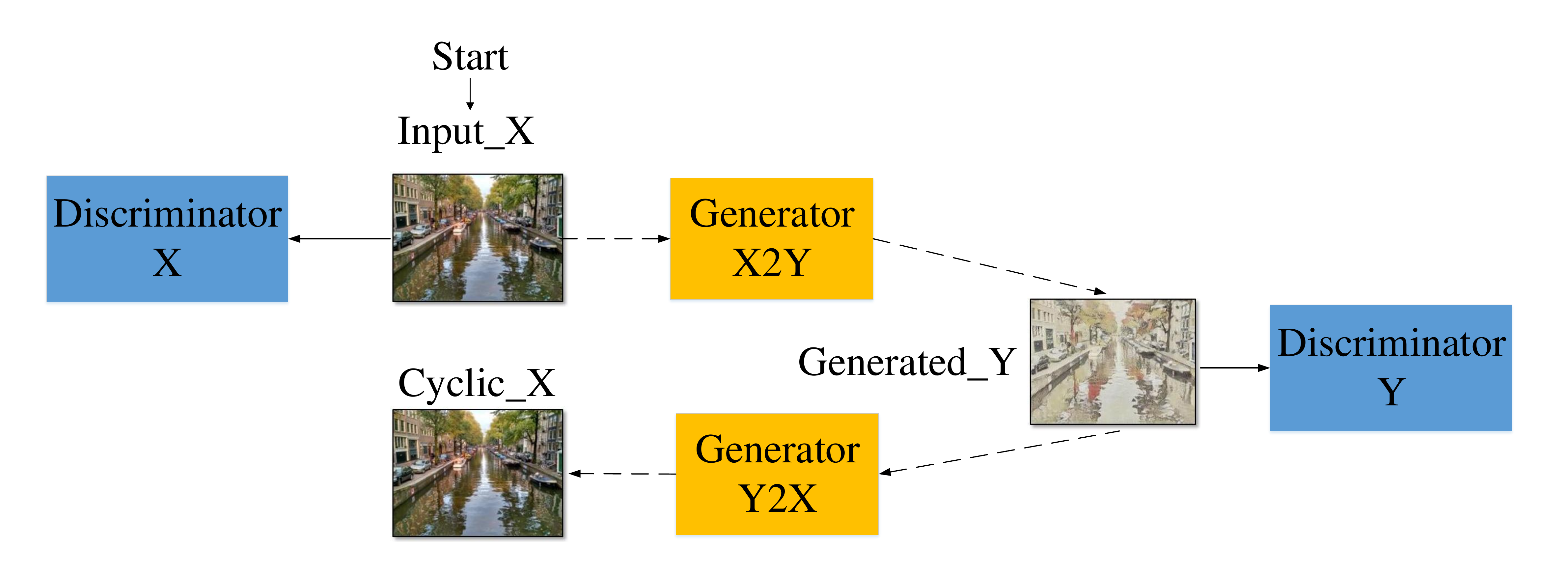}}
   \subfigure[]{
   \label{figureb}
   \includegraphics[width=1\linewidth]{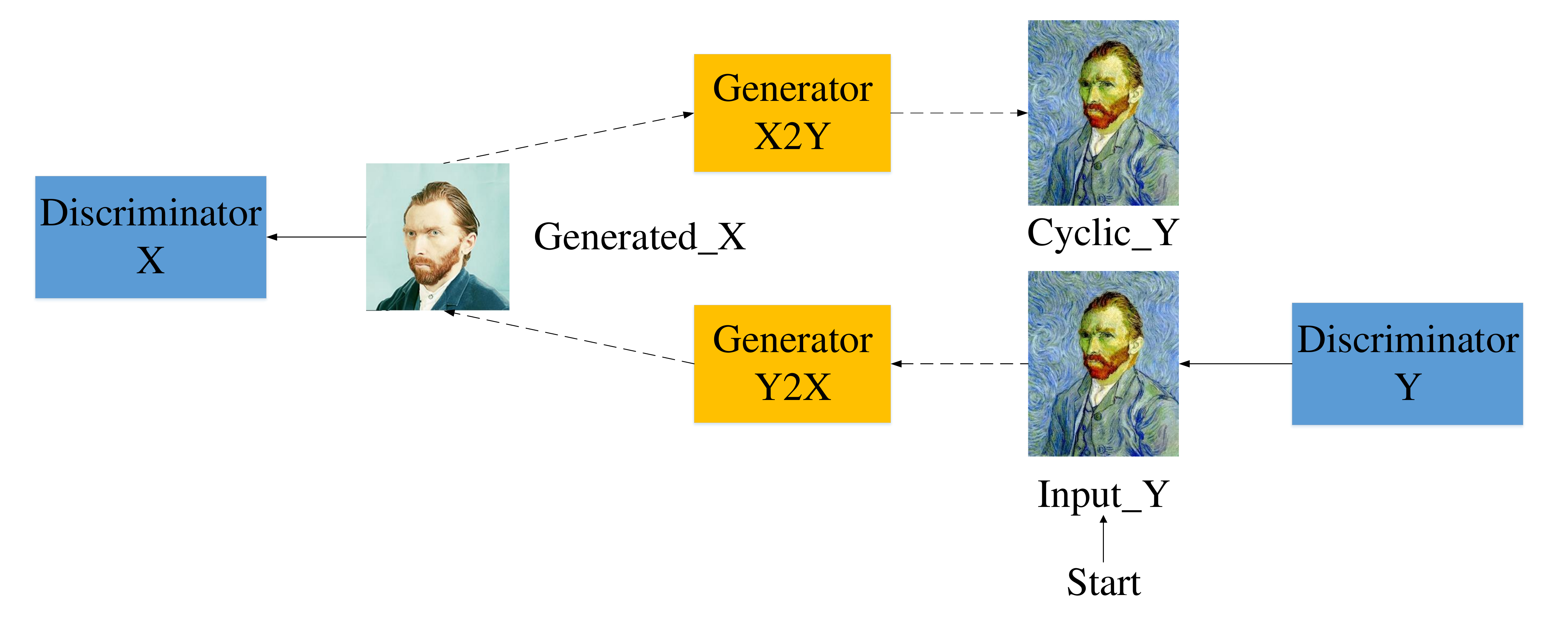}}
   \caption{An example demonstration of CycleGAN. (a) transform the real picture into Van Gogh style painting; (b) transform Van Gogh style painting into the real pictures.} \label{CycleGAN}
\end{figure}

Adversarial Example Generator (\emph{AEG}) is an important part of our approach. To improve the stability and security for target DL systems, \emph{AEG} provides effective adversarial examples to detect potential defects. The idea of generating adversarial examples is to add perturbations that people cannot distinguish from the original examples; this is very similar to the idea of GAN~\cite{Goodfellow2014Generative} generation of examples. GAN's generators $G$ and discriminators $D$ alternately generate adversarial examples that are very similar but not identical to the original examples based on noise data. Considering the difference of datasets of different target DL systems, such as some DL systems with label data and other DL systems may not, we choose \emph{CycleGAN}~\cite{Zhu2017Unpaired} as the training model of adversarial example generator, since \emph{CycleGAN} does not require the matching of data sets and label information.  \emph{CycleGAN} 
 is one of the most effective adversarial generation approaches. The mapping function and loss function of \emph{CycleGAN} are described as follows.
\begin{itemize}
\item The goal of \emph{CycleGAN} is to learn the mapping functions between two domains $X$ and $Y$. There are two mappings $G$ and $F$ in the model. There are two adversarial discriminators $Dx$ and $Dy$, where $Dx$ aims to distinguish between images $\{x\}$ and translated images $\{F(x)\}$. $Dy$ has a similar definition.
  \item Like other GANs, the adversarial loss function is used to optimize the mapping function. But during the actual training stage, it is found that the negative log- likelihood objective is not very stable and the loss function is changed to least-squares loss~\cite{Mao2016Least}.
 \item Because of the group mapping, it is impossible to train by using the adversarial loss function only. The reason is that the mapping $F$ can map all $x$ to a picture in $Y$ space, consequently, \emph{CycleGAN} puts forward the cycle consistency loss.
 \end{itemize}
 Fig.~\ref{CycleGAN} shows an example structure of \emph{CycleGAN}~\cite{Zhu2017Unpaired}. The purpose of this example is to transform real pictures and Van Gogh style paintings into each other. It does not need pairs of data to guide the adversarial  generation, and has a wide range and practicability. Therefore, in this paper, we use \emph{CycleGAN} to train our adversarial example generator, which can effectively generate adversarial examples to test the target DL systems.

\subsection{VGG-19 Network Structure}\label{subsec_featureEx}

\begin{figure}[h]
 \centering
 \includegraphics[width=\linewidth]{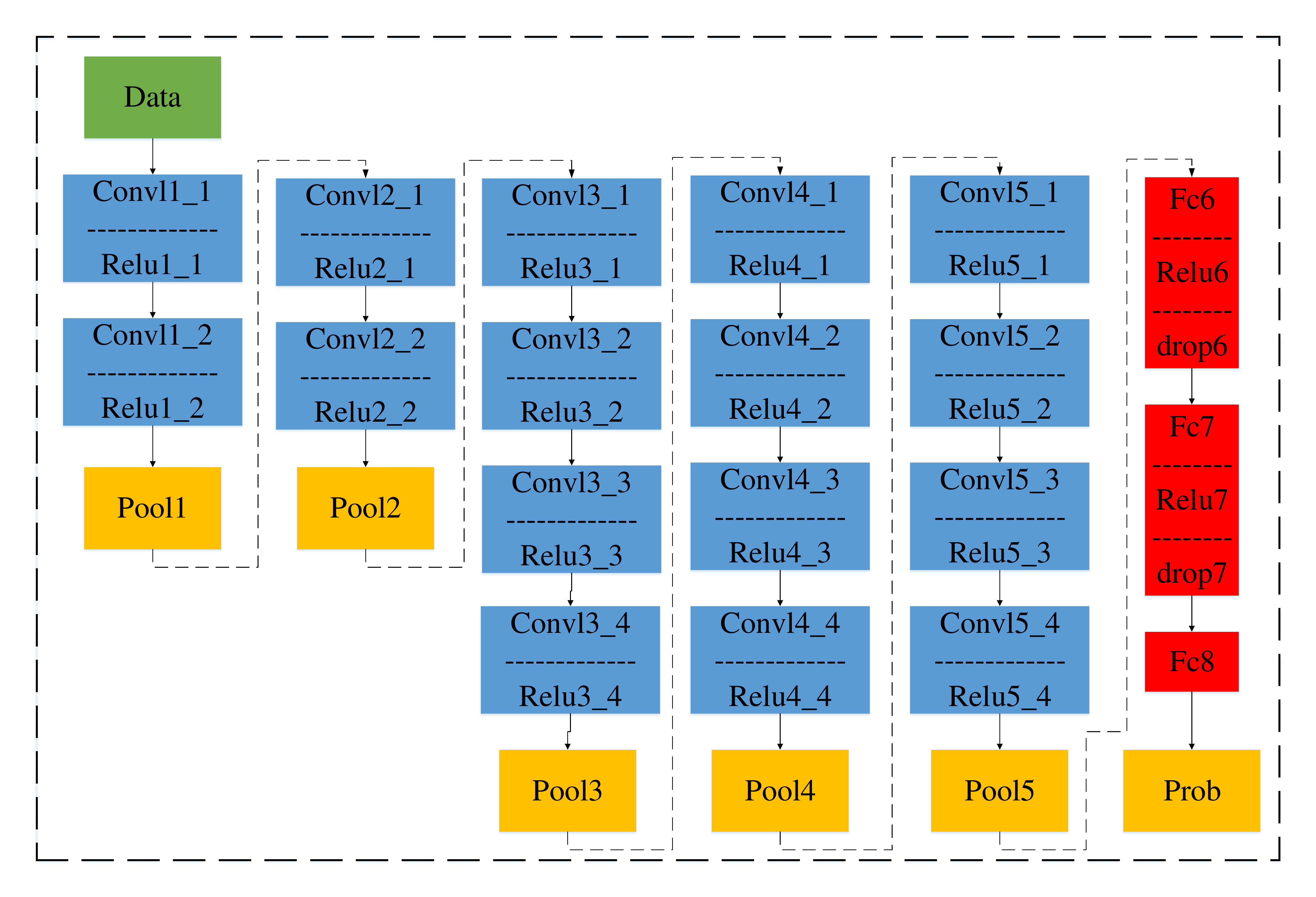}\\
 \caption{Structural Chart of VGG19 Network for Extracting Deep Features of Target Images}\label{VGG19}
\end{figure}

The ability of deep feature recognition and semantic expression extracted by CNN is stronger. Consequently, it has more advantages than traditional image features. The structure of VGG-19~\cite{Simonyan2014Very} convolution network is shown in Fig.~\ref{VGG19}. There are 19 layers including 16 convolution layers, i.e., two every Convl1-Convl2, four every Convl3-Convl5, and three full-connection layers, Fc6, Fc7, and Fc8. The works in~\cite{Donahue2014DeCAF,Babenko2014Neural} show that the VGG-19 network can extract high-level semantic information from images, and it can be used to identify similarities between images. In this paper, the output of the last full connection layer is fused as feature vector to compare the similarity between the adversarial examples and the original examples, and to serve as the threshold for filtering the generated adversarial examples.

\subsection{Neuron Coverage}\label{subsec_NC}
Pei et al.~\cite{Pei2017DeepXplore} propose for the first time neuron coverage as a measure of testing DL. They define neuron coverage of a set of test inputs as the ratio of the number of unique activated neurons in all test inputs to the total number of neurons in the DNN.


Let $N=\{n_{1},n_{2},...,n_{p}\}$ be the set of all neurons in the DNN, where $p$ is the length of neurons. The input to a DNN is an image $x_{i}\in T=\{x_{1},x_{2},...,x_{q}\}$, where $T$ is the input domain and $q$ is the length of the input domain.
Let $out (n_{i}, x_{i})$ be an output function that returns the output value of a neuron $n_{i}$ in DNN for a given test input $x_{i}$.
Finally, let $t$ represent the threshold for considering a neuron to be activated.
Then, the neuron coverage can be defined in the following:

\begin{equation}\label{NC}
\begin{split}
NC(T,x)=\frac{|\{n|\forall x\in T,out(n,x)> t\}|}{|N|}
\end{split}
\end{equation}

\section{Coverage-Guided Adversarial Generative Fuzzing Testing Approach}\label{sec_approach}
In this section, we first give an overview of our approach (Section~\ref{subsec_approachoverview}), and then we describe the pre-treatment
of our approach in Section~\ref{subsec_dataCollection}, including data collection and \emph{AEG} training.
Section~\ref{subsec_AC_gener} describes the algorithm of the adversarial example generation process. 
Finally, Section~\ref{subsec_coverage-guided} shows how our approach uses neuron coverage feedback to guide the generation of new adversarial examples.

\begin{figure}[tb]
 \centering
 \includegraphics[width=\columnwidth]{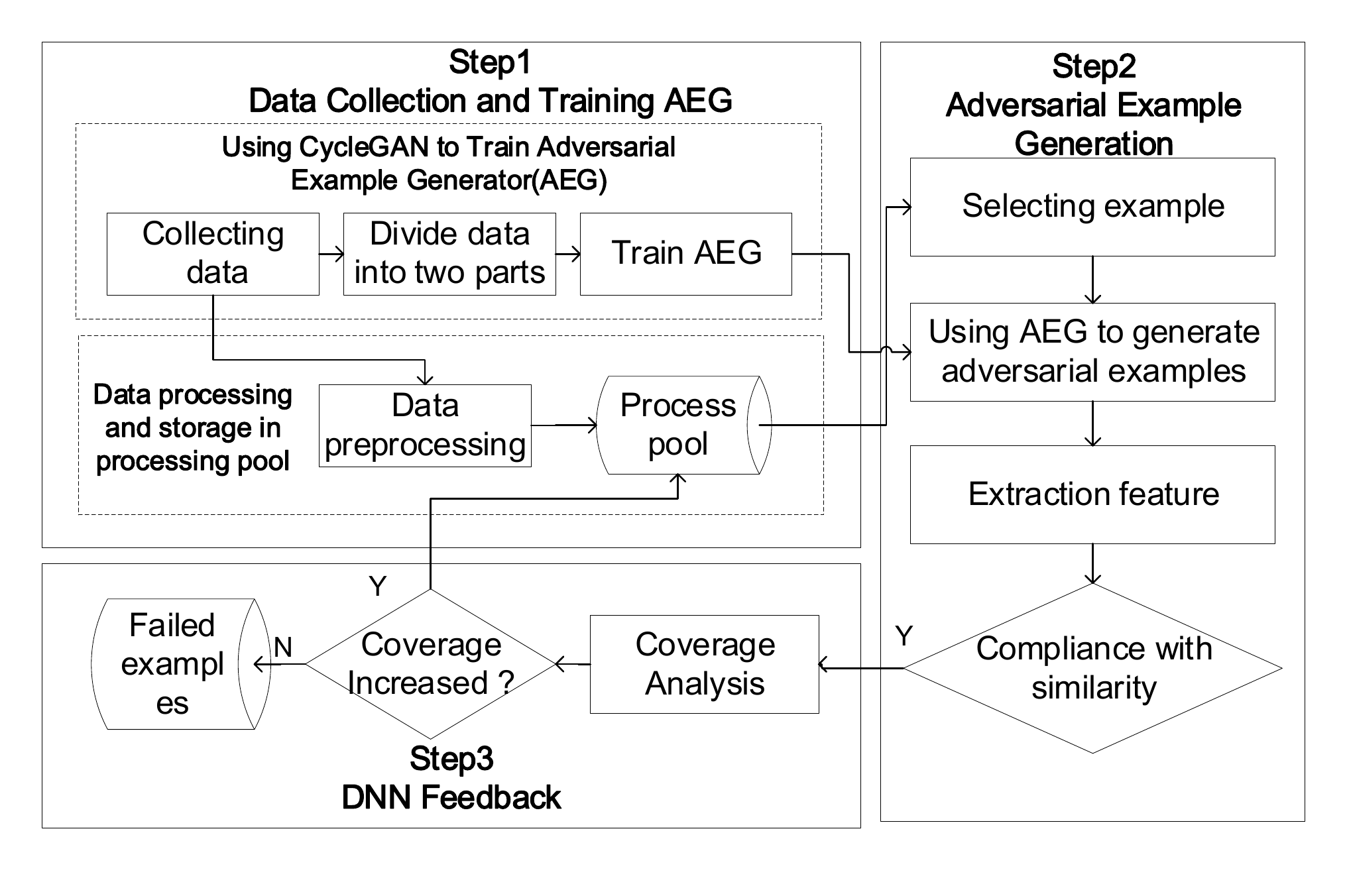}\\
 \caption{Coverage-Guided Adversarial Generative Fuzzing Testing Approach}\label{Fig2}
\end{figure}
\subsection{Overview}\label{subsec_approachoverview}

The core component of DL systems is the Deep Neural Network (DNN) with different structures and parameters. In the following discussions, we will study how to test DNNs. The input formats of DNNs can be various. In this paper, we focus on DNNs that take pictures as input. Adding perturbations to images has a great impact on DNNs and may cause errors. Guided by neuron coverage, the quality of the generated adversarial examples can be improved. As anticipated before, this paper presents \emph{CAGFuzz}, a coverage-guided adversarial generative fuzzing testing approach.
This approach generates adversarial examples with invisible perturbations based on \emph{AEG}. In general, Fig.~\ref{Fig2} shows the main process of our approach, which consists of three parts, described as follows:
\begin{itemize}
 \item The \emph{first step} is the data collection and training adversarial example generator. 
 For each data set, 
 the data sets are divided into two subsets and as the input of \emph{CycleGAN} to train \emph{AEG}. 
 These examples are then put into the processing pool after priority is set according to storage time. 
 We use this processing pool as the initial input for fuzzy testing.

 \item The \emph{second step} is the adversarial example generation. Each time a prioritized raw example is selected from the processing pool and used as the input of \emph{AEG} to generate adversarial examples. Using deep feature constraint to determine which adversarial examples should be saved. First, we use the VGG-19 network to extract the deep features (see Section~\ref{deep_feature}) of the original and adversarial examples. Then, we calculate the cosine similarity (see Section~\ref{cosine_similarity}) between the deep features of the original and the adversarial examples. If the cosine similarity between the two deep features is more than 0.9, we assume that the adversarial example is consistent with the original example in deep semantics and can be saved.

 \item The \emph{third step} is to use neuron coverage to guide the generation process. The adversarial examples generated in the second step is given as input to the DNN under test for coverage analysis. If a new coverage occurs, the adversarial example will be put into the processing pool as part of the dataset. The new coverage means that the neuron coverage of the adversarial example is higher than the neuron coverage of the original example.

\end{itemize}

\begin{algorithm}[h]
\caption{A description of the main loop of \emph{CAGFuzz}}
\begin{algorithmic}[1]
\REQUIRE $D$: Corresponding data sets, \\\qquad$DNN$: Target Deep Neural Network, \\\qquad$N$: The number of maximum iteration, \\\qquad$N1$: Number of new examples generated, \\\qquad$K$: Top-k parameter
\ENSURE Test Example Set for Increasing Coverage
\STATE $X,Y$ = Divide($D$);\label{L1}
\STATE Train \emph{AEG} through $X$ and $Y$\label{L2}
\STATE $T$ = Preprocessing($D$);\label{L3}
\STATE The preprocessed dataset $T$ serves as the initial processing pool\label{L4}
\WHILE{$number$ $of$ $iterations < N$}\label{L5}
\STATE $S$ = HeuristicSelect($T$);\label{L6}
\STATE $parent$ = Sample($S$);\label{L7}
\WHILE{$number$ $of$ $generation < N1$}\label{L8}
\STATE $data$ = AEG($parent$);\label{L9}
\STATE $Fp,Fd$ = FeatureExtraction($parent$,$data$);\label{L10}
\STATE $Similarity$ = CosineSimilarity($Fp$,$Fd$);\label{L11}
\ENDWHILE\label{L12}
\STATE Selecting top-k examples from all new examples;\label{L13}
\WHILE{$number$ $of$ $calculation < K$}\label{L14}
\STATE $cov$ = DNNFeed($data$);\label{L15}
\IF{IsNewCoverage($cov$)}\label{L16}
\STATE Add $data$ to processing pool \label{L17}
\STATE Setting time priority for $data$;\label{L18}
\ENDIF\label{L19}
\ENDWHILE
\ENDWHILE
\STATE Output all examples in the processing pool as a test example set;\label{L22}
\end{algorithmic}\label{algorithm}
\end{algorithm}

The main flow chart of the \emph{CAGFuzz} approach is shown in Algorithm~\ref{algorithm}. The input of \emph{CAGFuzz} includes a target Dataset ($D$), a deep neural network DNN ($DNN$), the number of maximum iterations $N$, the number of adversarial examples $N1$ generated by each original example, and the parameter $K$ of top-$k$. The output is the generated test example that improves the coverage of the target DNN.

Before the whole fuzzing process, we need to process the dataset. On the one hand, the dataset is divided into two equal data fields (Line~\ref{L1}) to train adversarial example generator \emph{AEG}(Line~\ref{L2}). On the other hand, all examples are pre-processed (Line~\ref{L3}) and stored in the processing pool (Line~\ref{L4}). During each iteration process (Line~\ref{L5}), the original example $parent$ is selected from the processing pool according to the time priority (Lines~\ref{L6}-~\ref{L7}). Then, each original example $parent$ is generated many times (Line~\ref{L8}). For each generation, the adversarial example generator \emph{AEG} is used to mutate the original example $parent$ to generate the adversarial example $data$ (Line~\ref{L9}). The deep features of the original example $parent$ and the adversarial example $data$ are extracted separately, and the cosine similarity (Lines~\ref{L10}-~\ref{L11}) between them is calculated.
Finally, all the adversarial examples generated by original example are sorted from high to low in similarity, and top-$k$ of them are selected as the target examples (Line~\ref{L13}). Calculating the neuron coverage of the top-$k$ adversarial examples and feedback these coverage to determine whether the adversarial example is saved (Line~\ref{L15}).
If the adversarial examples increase the coverage of the target DNN, they will be stored in the processing pool and set a time priority (Lines~\ref{L16}-~\ref{L19}).
The content of time priority is in Section~\ref{Example_prio}.

\subsection{Data Collection and Training AEG} \label{subsec_dataCollection}
\subsubsection{Data Collection}\label{dataColl}
We define the target task of \emph{CAGFuzz} as an image classification problem. Image classification is the core module of most existing DL systems. The first step of \emph{CAGFuzz} is to choose the image classification DNN (e.g. LeNet-1, 4, 5) to be tested and the dataset to be classified. The operation of the dataset is divided into two parts. First, all the examples in the dataset are prioritized, and then all the examples are stored in the processing pool as the original example. During the process of fuzzing, the fuzzer selects the original example from the processing pool according to the priority to perform the fuzzing operation. Second, the dataset is divided into two uniform groups. According to the domain, it is used as the input of the cycle generative adversarial network to train the adversarial example generator.

\subsubsection{Training Adversarial Example Generator}\label{train_AEG}
Traditional fuzzers mutate the original examples by flipping bits/bytes, cross-input files and swap blocks to achieve the effect of fuzziness. However, mutation of DNN input using these methods is not achievable or invalid, and may produce a large number of invalid and/or non-semantic testing examples.
At the same time, how to grasp the degree of mutation is also a question for us to think about. If mutation changes very little, the newly generated examples may be almost unchanged. Although this may be meaningful, the possibility of new examples finding DNN errors is very low. On the other hand, if the mutation changes greatly, more defects of DNN may be found. However, the semantics gap between the new generated example and the original example may be also large, that is to say, the new generated example is also invalid.

We propose a new strategy that uses adversarial example generator as mutations.
Given an image example $x$, \emph{AEG} generates an adversarial example $x'$, and the deep semantics information of $x'$ is consistent with that of $x$, but the adversarial perturbations that cannot be observed by human eyes are added. We invert the idea of \emph{CycleGAN}, add adversarial perturbations to the original example by adversarial loss, and control the perturbations to be invisible to human eyes by cyclic consistency loss.

\begin{figure}[t]
 \centering
 \includegraphics[width=0.8\linewidth]{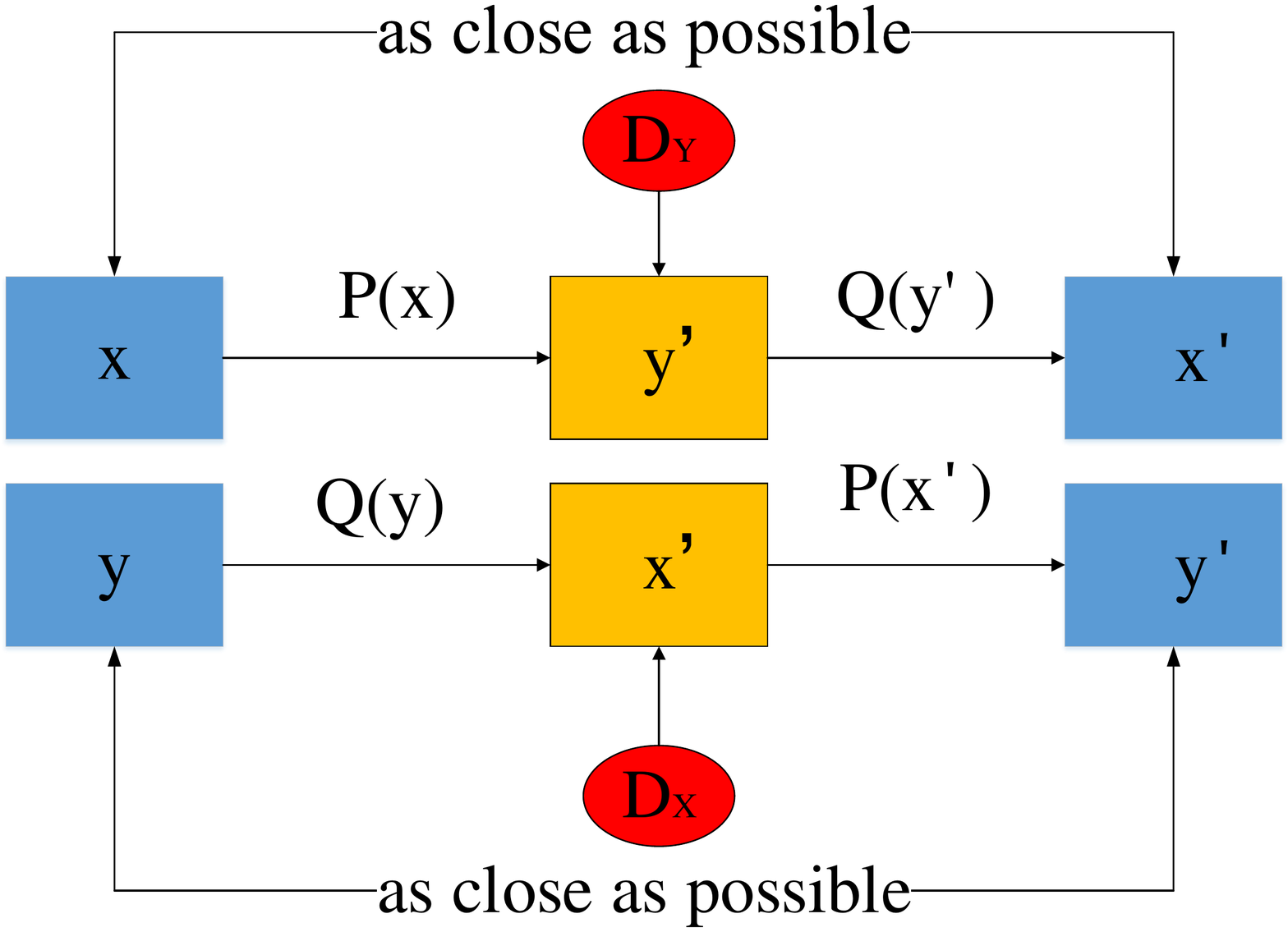}\\
 \caption{Transformation relationship between two mapping functions in training \emph{AEG}.}\label{Fig4}
\end{figure}

In Section~\ref{dataColl}, we propose to divide the collected data into two groups of data domains evenly. We define these two data domains as data domain $X$ and data domain $Y$. Our goal is to use the two data domains as input of the \emph{CycleGAN}, and to learn mapping functions from each other between the two data domains to train the \emph{AEG}. Supposing that the set of data domain $X$ is represented as $\{x_{1}, x_{2},..., x_{n}\}$, where $x_{i}$ denotes a training example in data domain $X$. Similarly, the set of data domain $Y$ denotes $\{y_{1}, y_{2},..., y_{m}\}$, where $y_{i}$ represents a training example in data domain $Y$. We define the data distribution of two groups of data domains, where the data domain $X$ is expressed as $x\sim P_{data}(x)$, and data domain $Y$ is expressed as $y\sim P_{data}(y)$. As shown in Fig.~\ref{Fig4}, the mapping functions between two sets of data domains are defined as $P:X\rightarrow Y$ and $Q:Y\rightarrow X$, where $P$ represents the transformation from data domain $X$ to data domain $Y$, and $Q$ represents the transformation from data domain $Y$ to data domain $X$. In addition, there are two adversarial discriminators $D_{X}$ and $D_{Y}$. $D_{X}$ distinguishes the original example ${x}$ of data domain $X$ from the one generated by mapping function $Q$. Similarly, $D_{Y}$ distinguishes the original example ${y}$ of data domain $Y$ from the adversarial example ${P(x)}$ generated by mapping function $P$.

\emph{Adversarial Loss.} The mapping function between two sets of data domains is designed with loss function. For mapping function $P$ and corresponding adversarial discriminator $D_{Y}$, the objective function is defined as follows:
\begin{equation}\label{T}
\begin{split}
\min_P\max_DY V(P,D_{Y},X,Y)=E_{y\sim P_{data}(y)}[logD_{Y}(y)]+\\E_{x\sim P_{data}(x)}[log(1-D_{Y}(P(x)))]
\end{split}
\end{equation}

The function of mapping function $P$ is to generate adversarial examples $y'= P(x)$ similar to data domain $Y$, which can be understood as adding large perturbations with $Y$ characteristics of data domain to the original example $x$ of data domain $X$.
At the same time, there is an adversarial discriminator $D_{Y}$ to distinguish the real examples $y$ in data domain $Y$ and the generated adversarial example $Y'$. 
The objective of the objective function is to minimize the mapping function $P$ and maximize the adversarial discriminator $D_{Y}$.
Similarly, for the mapping function $Q$ and the target function set by the adversarial discriminator $D_{X}$, the objective function is defined in the following:
\begin{equation}\label{T}
\begin{split}
\min_Q\max_DX V(Q,D_{X},Y,X)=E_{x\sim P_{data}(x)}[logD_{X}(x)]+\\E_{y\sim P_{data}(y)}[log(1-D_{X}(Q(y)))]
\end{split}
\end{equation}

\emph{Cycle Consistency Loss.} We can add perturbations to the original example by using the aforementioned adversarial loss function, but the degree of mutation of this perturbation is large, and it is prone to generate
invalid adversarial examples. To avoid this problem, we add constraints to the perturbations, and control the degree of mutation through the cycle consistency loss. In this way, the perturbation-resistant human eyes added to the original example are invisible. For example, example $x$ of data domain $X$ is generated by mapping function $P$ to generate adversarial example $y'$, and then adversarial example $y'$ is generated by mapping function $Q$ to generate new adversarial example $x'$. At this time, the generated adversarial example $x'$ is similar to the original example $x$, that is to say,
$x \rightarrow P(x)=y' \rightarrow Q(y')=x' \approx x$.
The objective function of the loss function of cyclic consistency is described as follows:
\begin{equation}\label{T}
\begin{split}
Loss_{cycle}(P,Q)=E_{x\sim P_{data}(x)}[||Q(P(x)-x||_{1}]+\\E_{y\sim P_{data}(y)}[||P(Q(y)-y||_{1}]
\end{split}
\end{equation}

\begin{figure*}
\centering
   \subfigure[automobile and truck]{
   \label{figure5a}
   \includegraphics[width=0.32\linewidth]{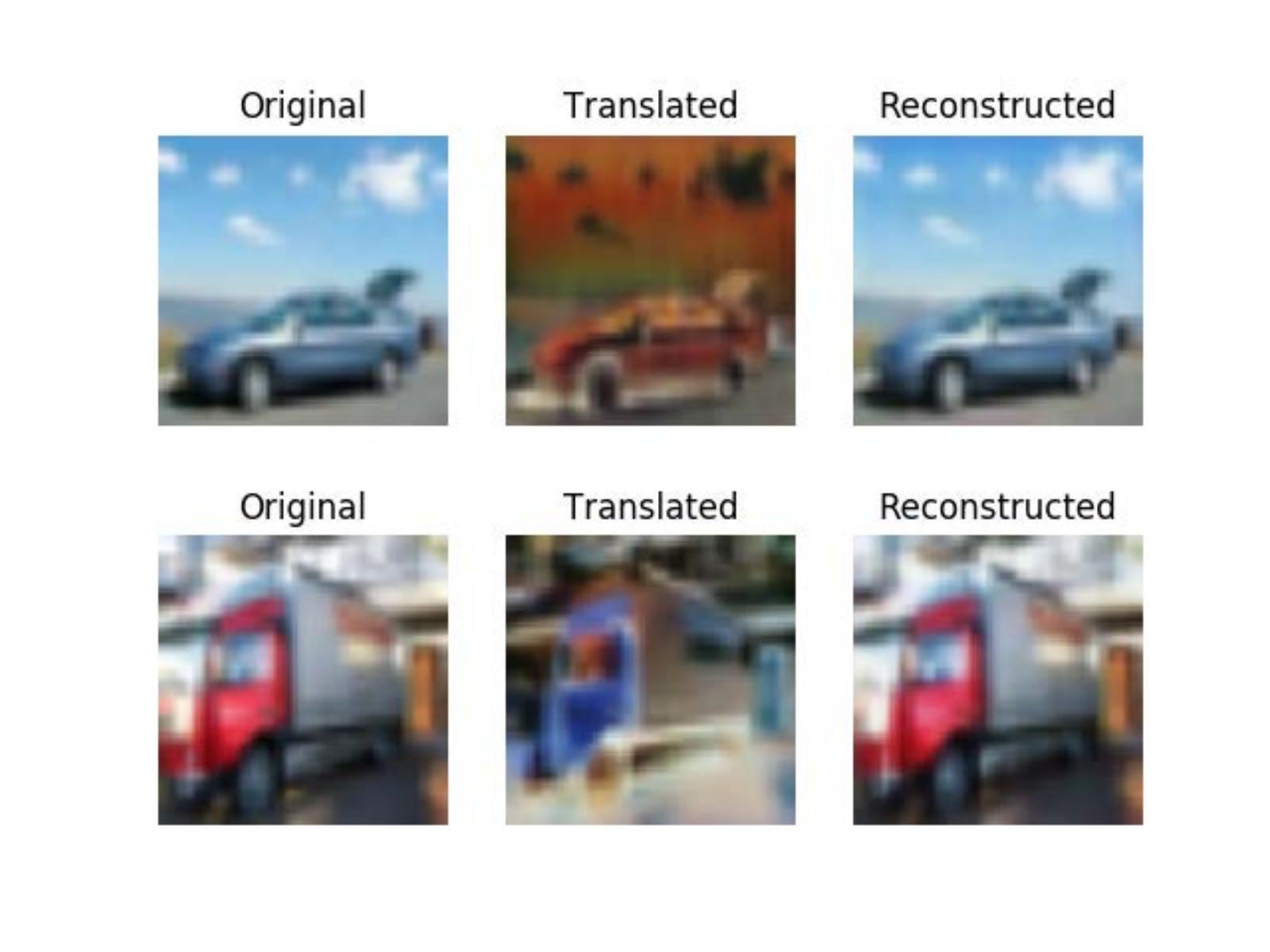}}
   \subfigure[airplane and bird]{
   \label{figure5b}
   \includegraphics[width=0.32\linewidth]{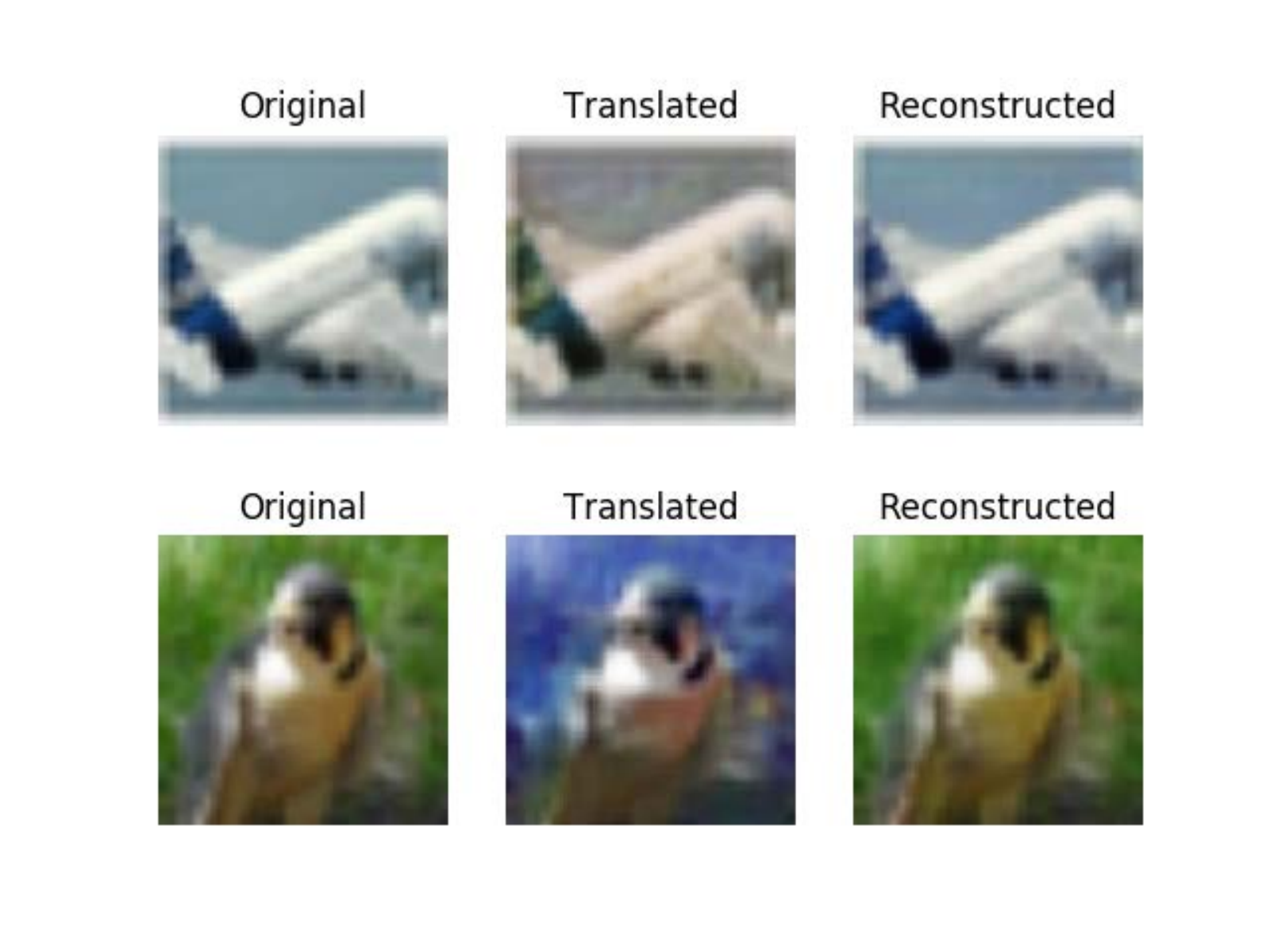}}
   \subfigure[frog and ship]{
   \label{figure5c}
   \includegraphics[width=0.32\linewidth]{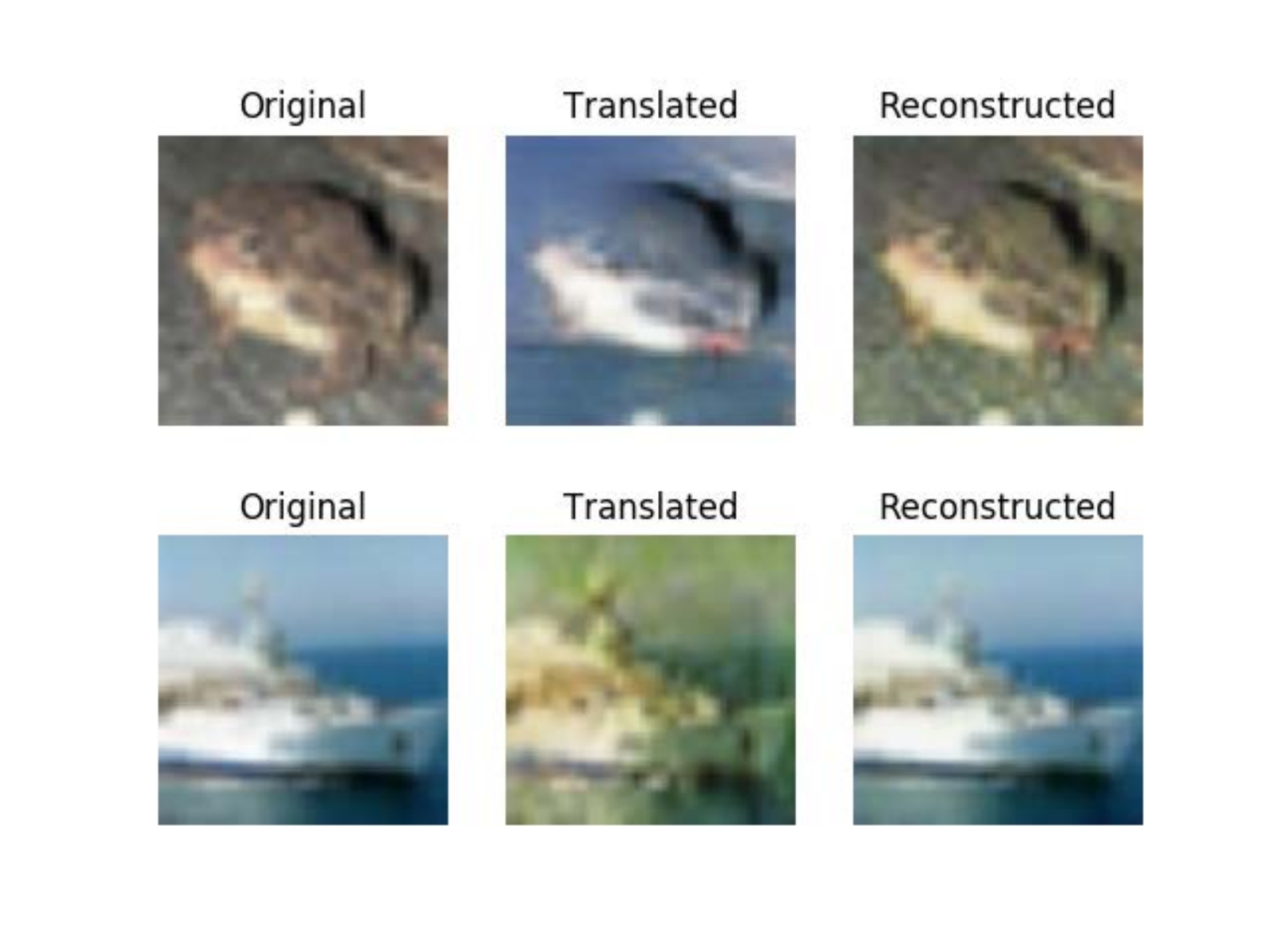}}
   \subfigure[horse and deer]{
   \label{figure5d}
   \includegraphics[width=0.32\linewidth]{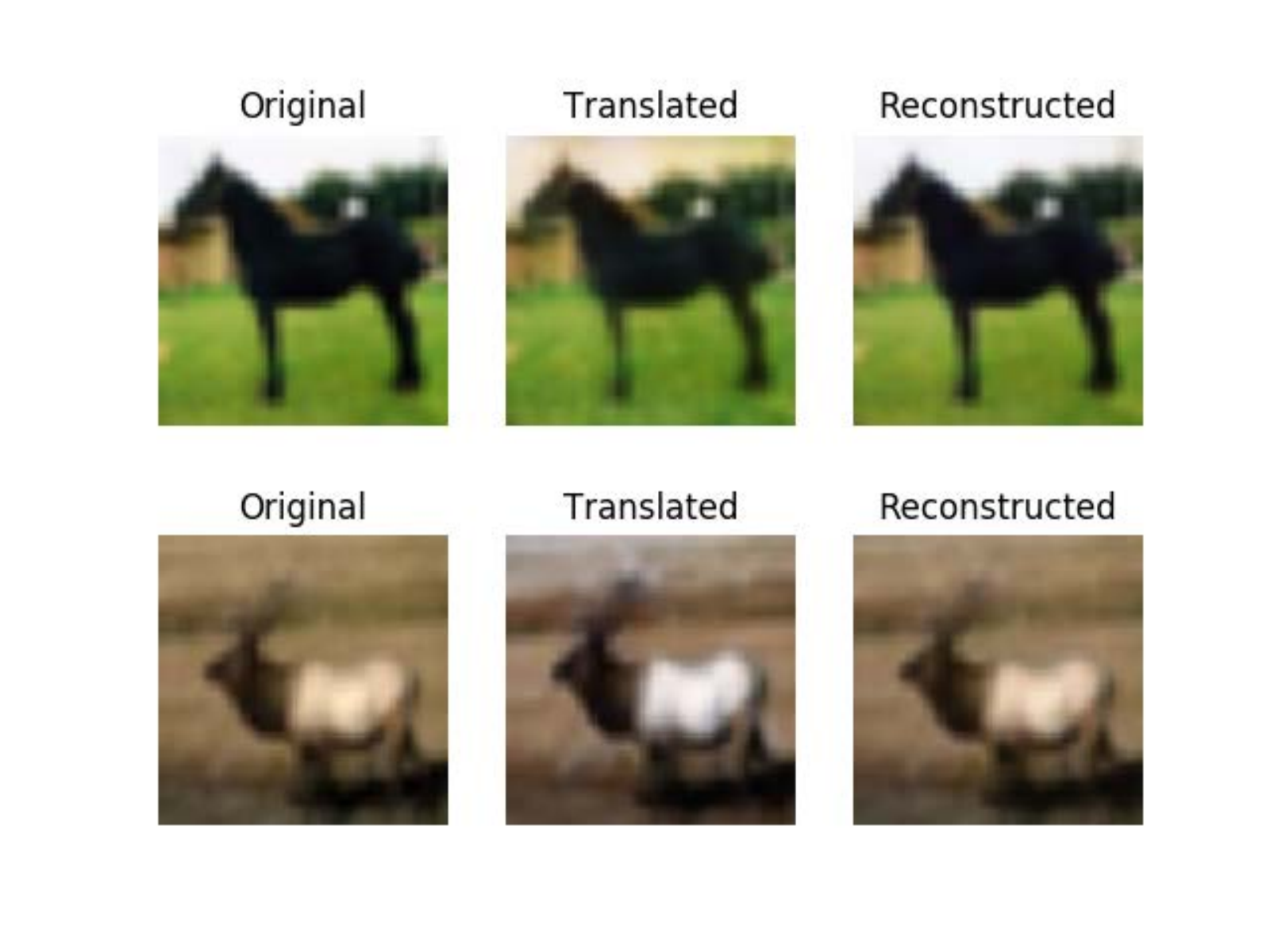}}
   \subfigure[dog and cat]{
   \label{figure5e}
   \includegraphics[width=0.32\linewidth]{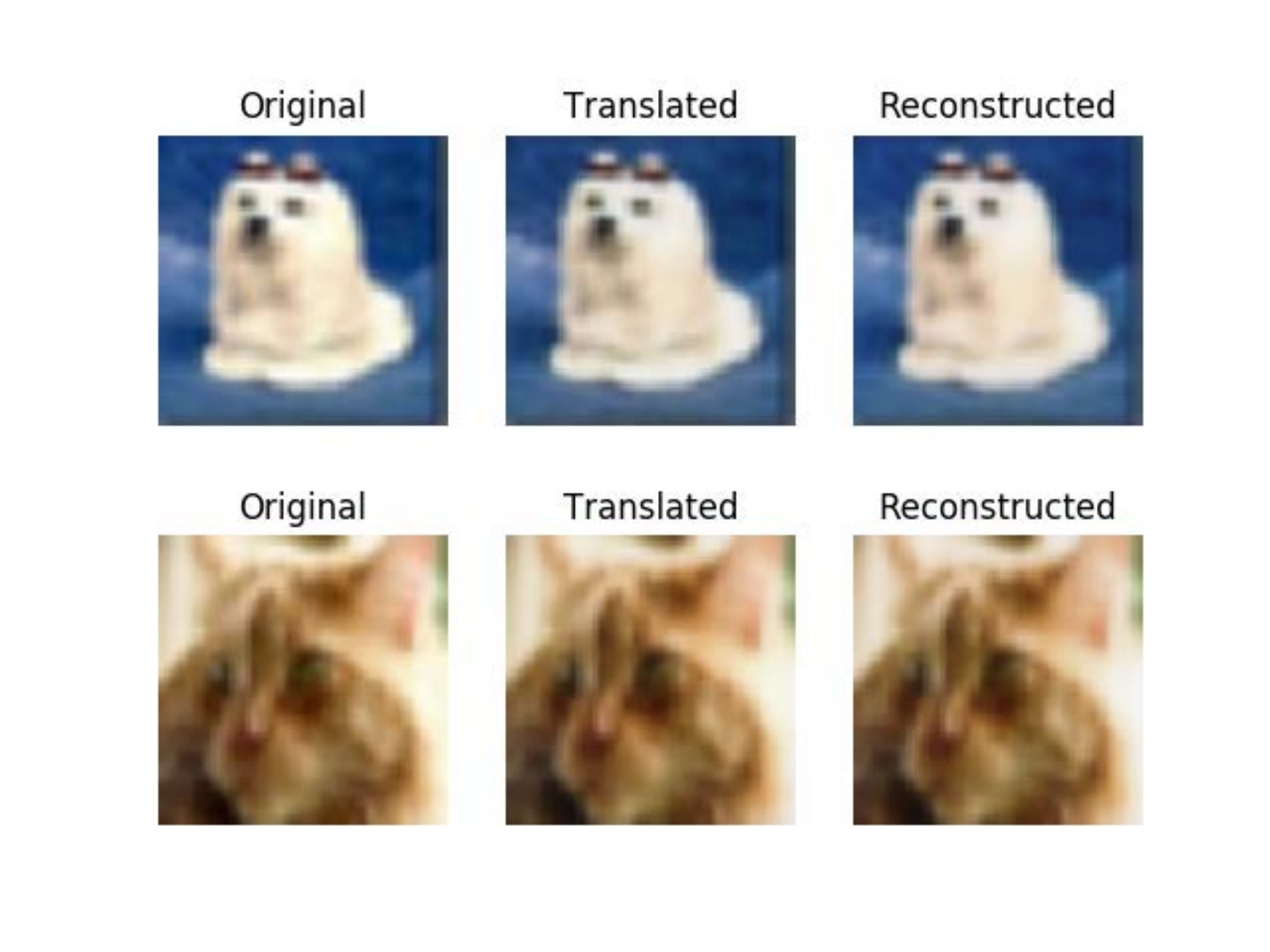}}
   \subfigure[apple and orange]{
   \label{figure5f}
   \includegraphics[width=0.32\linewidth]{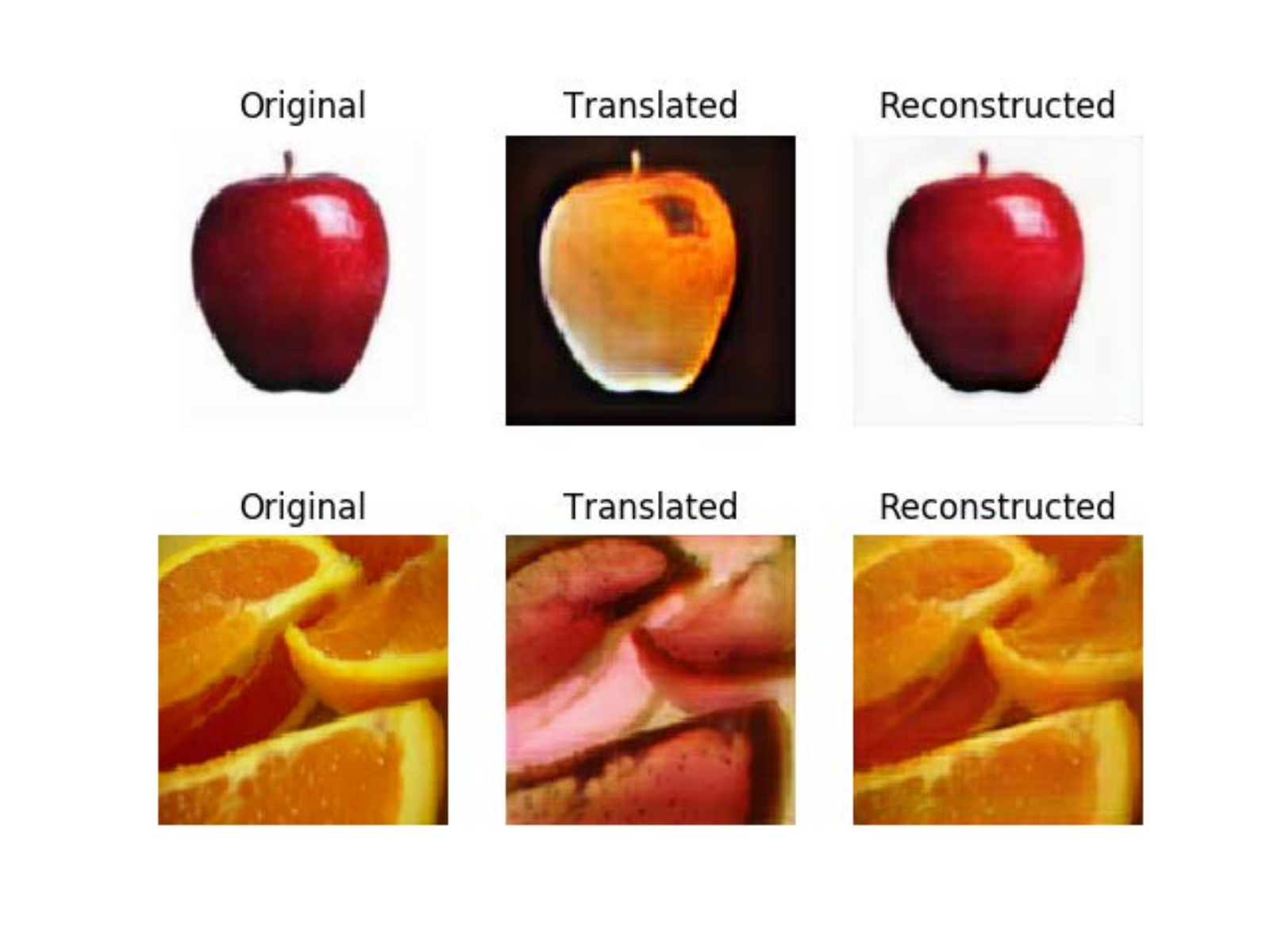}}
   \caption{\emph{AEG} generates effect maps of adversarial examples. In each picture, the leftmost column is the original example, the middle column is the transformed example of the original example, and the rightmost column is the reconstructed example.} \label{Fig3}
\end{figure*}

The overall structure of the network has two generators: $P$ and $Q$, and two discriminator networks $D_{X}$ and $D_{Y}$. The whole network is a dual structure. We combine two generators with opposite functions into our adversarial example generator. The effect picture of \emph{AEG} is shown in Fig.~\ref{Fig3}, we show that the adversarial example generation process has 12 groups of pictures of different categories. In each picture, the leftmost column is the original example, the middle column is the transformed example of the original example, and the rightmost column is the reconstructed example. We choose the reconstructed example as the adversarial example. First, larger perturbations are added to the original example. Second, the degree of mutation is controlled by reverse reconstruction to generate adversarial examples with smaller perturbations.

\subsection{Adversarial Example Generation} \label{subsec_AC_gener}

\subsubsection{Example Priority}\label{Example_prio}
The priority of the example determines which kind of examples should be selected next time. We choose a probabilistic selection strategy based on the time of adding examples to the processing pool. We adopt a meta-heuristic formula with faster selection speed. The probability calculation formula is described as follows: $h(b_{i}, t)=\frac{e^{t_{i}-t}}{\sum e^{t_{i}-t}}$, where $h(b_{i}, t)$ represents the probability of selecting example $b_{i}$ at time $t$, and $t_{i}$ represents the time when example $b_{i}$ joins the processing pool.

This priority can be understood as follows: the most recently sampled examples are more likely to generate useful new neuron coverage when mutating to adversarial examples. However, when time passes, the advantage will gradually diminish.

\subsubsection{Deep Feature}\label{deep_feature}
To ensure the meaning of the generated adversarial examples as much as possible, we adopt the strategy of extracting the semantics features of the original examples and adversarial examples and controlling their differences within a certain range. The deep feature recognition ability and semantics expression ability extracted by CNN are stronger.
Consequently, we select the VGG-19 network to extract the deep features of examples. The deep features in the VGG-19 model are extracted according to the hierarchy. Compared with the high-level features, the low-level features are unlikely to contain rich semantics information.

The deep features extracted from the VGG-19 network model can represent images better than traditional image features. It also shows that the deeper the layer of convolution network, the more parameters in the network, and the better the image can be expressed. We fuse the output of the last full connection layer (Fc8 layer in Fig.~\ref{VGG19}) as deep feature, and the dimension of deep feature is 4096.

\subsubsection{Cosine Similarity Computation}\label{cosine_similarity}

During the mutation process, \emph{AEG} generates multiple adversarial examples for each original example. We assume that the original example is $a$, and the set of all adversarial examples is $T=\{a_{1}, a_{2},...,a_{n}\}$, which extracts the semantics feature vectors for the original example and all the confrontational examples by the feature extraction method mentioned above. The dimension of each feature vector is 4096. Supposing that the feature vector corresponding to the original example $a$ is $X=[x_{1}, x_{2},...,x_{n}]_{n = 4096}$, and the corresponding eigenvector of an adversarial example is $a_{i}$ is $Y=[y_{1}, y_{2},...,y_{n}]_{n = 4096}$, where $a_{i}\in T$. Cosine similarity is used to measure the difference between each adversarial example and the original example. The formula is described as follows:
\begin{equation}\label{T}
\begin{split}
COS(X,Y)=\frac{X\cdot Y}{||X||\times ||Y||}=\frac{\sum_{i=1}^{n}(x_{i}\times y_{i})}{\sqrt{\sum_{i=1}^{n}x_{i}^{2}}\times \sqrt{\sum_{i=1}^{n}y_{i}^{2}}}
\end{split}
\end{equation}
where $x_{i}$ and $y_{i}$ correspond to each dimension of eigenvector $X$ and $Y$.

To control the size and improve the mutation quality of adversarial examples, we select the top-$k$ adversarial examples sorted from high to low cosine similarity as eligible examples to continue the follow-up steps. In our approach, we set $K=5$, that is to say, we select the five adversarial examples with the highest cosine similarity for neuron coverage.

\subsection{DNN Feedback}\label{subsec_coverage-guided}
Without coverage as a guiding condition, the adversarial examples generated by \emph{AEG} are not purposeful. Consequently it is impossible to know whether the adversarial examples are effective or not. If the generated adversarial examples cannot bring new coverage information to the DNN to be tested, these adversarial examples can only simply expand the dataset, but cannot effectively detect the potential defects of DNN. To make matters worse, mutations in these adversarial examples may bury other meaningful examples in a fuzzy queue, thus significantly reducing the fuzzing effect. Therefore, neuron coverage feedback is used to determine whether the newly generated adversarial examples should be placed in the processing pool for further mutation.

After each round of generation and similarity screening, all valid adversarial examples are used as the input of DNN to be tested for neuron coverage analysis. If the adversarial examples generate new neuron coverage information, we will set priority for the adversarial examples and store it in the processing pool for further mutation.
For example, a DNN for image classification consists of 100 neurons. 32 neurons are activated when the original example is input into the network, and 35 neurons are activated when the adversarial example is input into the network. Consequently, we say that the adversarial example brings new coverage information.

\section{Experimental Evaluation}\label{sec_validation}
In this section, we perform a set of dedicated experiments to validate
\emph{CAGFuzz}. Section~\ref{subsec_question} proposes the research questions.
Section~\ref{subsec_design} describes the experimental design.
Section~\ref{subsec_result} provides the experimental results and
Section~\ref{subsec_Threats} discusses some threats to validity.
\subsection{Research Questions}\label{subsec_question}
We use three standard deep learning datasets and the corresponding image classification models to carry out a series of experiments to validate \emph{CAGFuzz}. The purpose of the experiments is designed to explore the following four main research questions:

\begin{itemize}
    \item  \emph{RQ1}: Could the generated adversarial examples based on data have stronger generalization ability than those based on models?

    \item  \emph{RQ2}: Could \emph{CAGFuzz} improve the neuron coverage in the target network?

    \item  \emph{RQ3}: Could \emph{CAGFuzz} find potential defects in the target network?


    \item  \emph{RQ4}: Could the accuracy and the robustness of the target network be improved by adding adversarial examples to the training set?
\end{itemize}

To discover potential defects of target network and expand effective examples for data sets, the \emph{CAGFuzz} approach mainly generates adversarial examples for DNNs to be tested. Therefore, we designed \emph{RQ1} to explore whether the examples generated based on data have better generalization ability than those based on models.
For neuron coverage, we designed \emph{RQ2} to explore whether \emph{CAGFuzz} can effectively generate test examples with more coverage information for target DNNs. We designed \emph{RQ3} to study whether \emph{CAGFuzz} can discover more hidden defects in target DNNs. \emph{RQ4} is designed to explore whether adding the adversarial examples generated by \emph{CAGFuzz} to the training set can significantly improve the accuracy of target DNNs.

\subsection{Experimental Design}\label{subsec_design}
\subsubsection{Experimental Environment}
The experiments have been performed on Linux machines. The detailed descriptions of the hardware and software environments of the experiments are
shown in Table~\ref{Tab1}.

\begin{table}[ht]
\caption{Experimental hardware and software environment}\label{Tab1}
\centering
\renewcommand\arraystretch{1.5}
\begin{tabular}{p{4cm}<{\centering} p{4cm}<{\centering} }
  \toprule
    Name & Standard \\ \hline
    CPU     & Xeon Silver 4108  \\
    GPU     & NVIDIA Quadro P4000  \\
    RAM     & 32G  \\
    System     & Ubuntu 16.04  \\
    Programming environment     & Python  \\
    Deep learning open source framework     & Tensorflow1.12  \\
  \toprule
\end{tabular}
\end{table}

\subsubsection{DataSets and Corresponding DNN Models}
For research purpose, we adopt three popular and commonly used datasets with different types of data: MNIST~\cite{Deng2012The}, CIFAR-10~\cite{Li2017CIFAR10}, and ImageNet~\cite{Russakovsky2015ImageNet}.
At the same time, we have learned and trained several popular DNN models for each dataset, which have been widely used by scientific researchers. In Table~\ref{Tab2}, we provide an informative summary of these datasets and the corresponding DNN models. All the evaluated DNN models are either pre-trained (i.e., we use the common weights in previous researchers' papers) or trained according to standards by using common datasets and public network structures.

MNIST~\cite{Deng2012The} is a large handwritten digital dataset containing $28 * 28 * 1$ pixels of images with class labels ranging from 0 to 9. The dataset contains 60,000 training examples and 10,000 test examples.
We construct three different kinds of neural networks based on LeNet family, namely LeNet-1, LeNet-4, and LeNet-5.

CIFAR-10~\cite{Li2017CIFAR10} is a set of general image classification images, including $32 * 32 * 3$ pixel three-channel images, including ten different kinds of pictures (such as aircraft, cats, trucks, etc.). The dataset contains 50,000 training examples and 10,000 test examples. Due to the large amount of data and high complexity of CIFAR-10, its classification task is more difficult than MNIST. To obtain the competitive performance of CIFAR-10, we choose three famous DNN models VGG-16, VGG-19, and ResNet-20 as the targeted models.

ImageNet~\cite{Russakovsky2015ImageNet} is a large image dataset, in which each image is a $224 * 224 * 3$ three-channel image, containing 1000 different types. The dataset contains a large number of training data (more than one million) and test data (about 50,000). Therefore, for any automated testing tool, working on ImageNet-sized datasets and DNN models is a severe test. Because the large number of images in the ImageNet dataset, most state-of-the-art adversarial approaches are only evaluated on a part of the ImageNet dataset.
To obtain the competitive performance of ImageNet, we choose three famous DNN models VGG-16, VGG-19, and ResNet-50 as the targeted models.

\begin{table}[h]
\caption{Subject datasets and DNN models}
\centering
\renewcommand{\multirowsetup}{\centering}
\begin{tabular}{p{1.2cm}<{\centering} p{1.5cm}<{\centering} p{1.3cm}<{\centering} p{0.6cm}<{\centering}p{0.8cm}<{\centering}p{.9cm}<{\centering}}
\toprule
\textbf{DataSet} &\textbf{DataSet Description} &\textbf{Model} &\textbf{\#Layer} &\textbf{\#Neuron} &\textbf{Test acc(\%)}\\
\midrule
\multirow{3}{*}{MNIST}& \multirow{3}{1.8cm}{Hand written digits from 0 to 9}
&LeNet-1 &7 &52  &98.25\\
& &LeNet-4 &8 &148 &98.75\\
& &LeNet-5 &9 &268 &98.63\\
\midrule
\multirow{3}{*}{CIFAR-10}& \multirow{3}{1.8cm}{10 class general image}
&VGG-16    &16 &19540 &86.84\\
&  &VGG-19    &19 &41118 &77.26\\
&  &ResNet-20 &70 &4861 &82.86\\
\midrule
\multirow{3}{*}{ImageNet}& \multirow{3}{1.8cm}{1000-class large scale datasets}
&VGG-16 &16 &14888   &92.6\\
& &VGG-19 &19 &16168  &92.7\\
& &ResNet-50  &176 &94059  &96.43\\
\bottomrule
\end{tabular}
\label{Tab2}
\end{table}


\subsubsection{Contrast Approaches}
As surveyed in~\cite{zhang2019machine}, there are several
open-source tools in testing machine learning applications. Some released tools, such as \emph{Themis}~\footnote{http://fairness.cs.umass.edu/}, \emph{mltest}~\footnote{ https://github.com/Thenerdstation/mltest}, and \emph{torchtes}~\footnote{https://github.com/suriyadeepan/torchtest} do not focus on generating adversarial examples.
Thus, to measure the ability of \emph{CAGFuzz}, we selected the following three representative DL testing approaches proposed recently in the literature as our contrast approaches, respectively:
 \begin{itemize}
  \item \emph{FGSM}~\cite{goodfellow2014explaining} (Fast Gradient Sign Method) - a typical approach generates adversarial examples based on model. Consequently, we use \emph{FGSM} to generate adversarial examples to compare with \emph{CAGFuzz}, and verify that the generated adversarial examples based on pure data have higher generalization ability than those based on models.
  \item \emph{DeepHunter}~\cite{Xie2018DeepHunter} - an automated fuzz testing framework for hunting potential defects of general-purpose DNNs. \emph{DeepHunter} performs metamorphic mutation to generate new semantically preserved tests, and leverages multiple plug-able coverage criteria as feedback to guide the test generation from different perspectives.
  \item \emph{DeepXplore}~\cite{Pei2017DeepXplore} - the first white box system for systematically testing DL systems and automatically identify erroneous behaviors without manual labels. \emph{DeepXplore} performs gradient ascent to solve a joint optimization problem that maximizes both neuron coverage and the number of potentially erroneous behaviors.
 \end{itemize}

Since there is no open source version of \emph{DeepHunter}~\cite{Xie2018DeepHunter}, we have implemented eight image transformation methods mentioned in \emph{DeepHunter}, and we use these eight methods to replace \emph{DeepHunter} for later experimental evaluation. The source code of \emph{FGSM} and \emph{DeepXplore} can be found on GitHub, and the tools are utilized for later experimental evaluation.
\subsection{Experimental Results}\label{subsec_result}

\subsubsection{Training of Target DNNs}\label{subsec_trainDNN}
To ensure the correctness and validate the evaluation results of the experiments, we carefully select several popular DNN models with competitive performance for each dataset. These DNN models have been proven to be standard in previous researchers' experiments. In our approach, we closely follow the common machine learning training practices and guidelines, and set the learning rate for training DNN model.
During the initialization process of DNN model learning rate,
if the learning rate is too high, the weight of the model will increase rapidly, which will have a negative impact on the training of the whole model. Consequently, the learning rate is set to a smaller value at the beginning. For the three LeNet models of the MNIST dataset, we set the learning rate as 0.05.

\begin{figure}[t]
 \centering
 \includegraphics[width=0.7\linewidth]{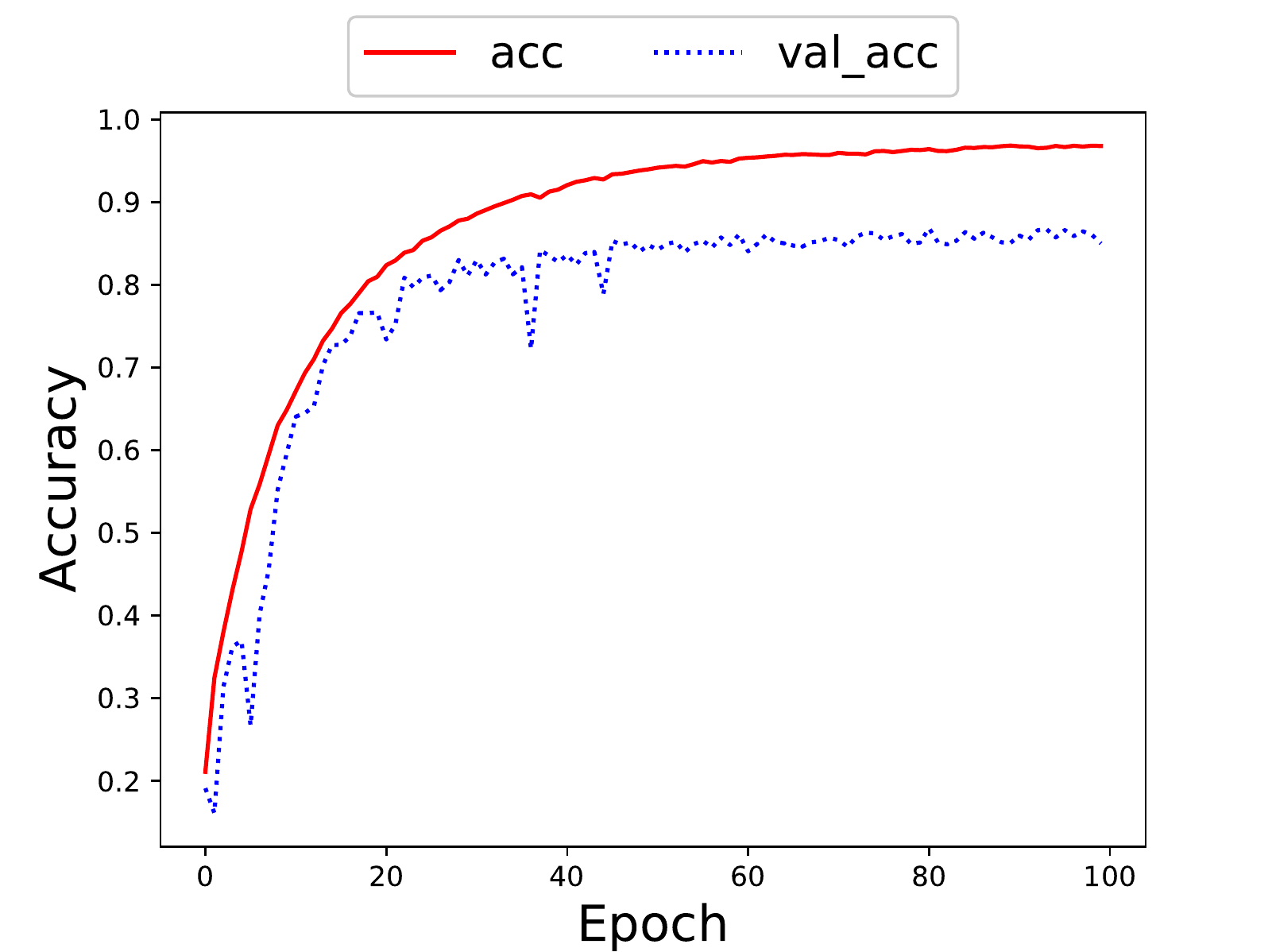}\\
 \caption{Sketch of VGG-16 network training accuracy and verification accuracy change with training epoch}\label{Fig5}
\end{figure}

For the two VGG networks of the CIFAR-10 dataset, we set the initial learning rate as 0.0005 based on experiences because of the deeper network layers and the more complex model.
In addition, we initially set the epoch for each model as 100 training times. The LeNet model works well, but when we train the VGG-16 network, we find that the accuracy of the model is basically stable after 50 training epochs, as shown in Fig.~\ref{Fig5}. Therefore, during the process of training the VGG-19 network and the subsequent retraining model stage, we reset the training epochs to 50; this can save a lot of computing resources and space-time costs.
During the process of training the ResNet-20 model, we set up a three-stage adaptive learning rate. When $epoch < 20$, we set the learning rate as $1e^{-3}$. When $20 < epoch < 50$, we set the learning rate as $1e^{-4}$. When $epoch > 50$, we set the learning rate as $1e^{-5}$.

For the VGG-16, VGG-19, and ResNet-50 models used to classify the ImageNet data sets, we directly used the model with ImageNet as the data set in the keras framework~\cite{ketkar2017introduction}, since it has already trained and achieved enough performance. The Imagenet data set is too large (including 137 GB training set, 12.7 GB test set, and 6.28 GB verification set), the cost of retraining model is too large, and the general hardware equipment cannot meet the requirements. Therefore, for the ImageNet data set, we only have performed experiments on the two modules (Neuron Coverage and Error Behavior Discovery).

\begin{figure*}
\centering
   \subfigure[LeNet-1]{
   \label{figure6a}
   \includegraphics[width=0.32\linewidth]{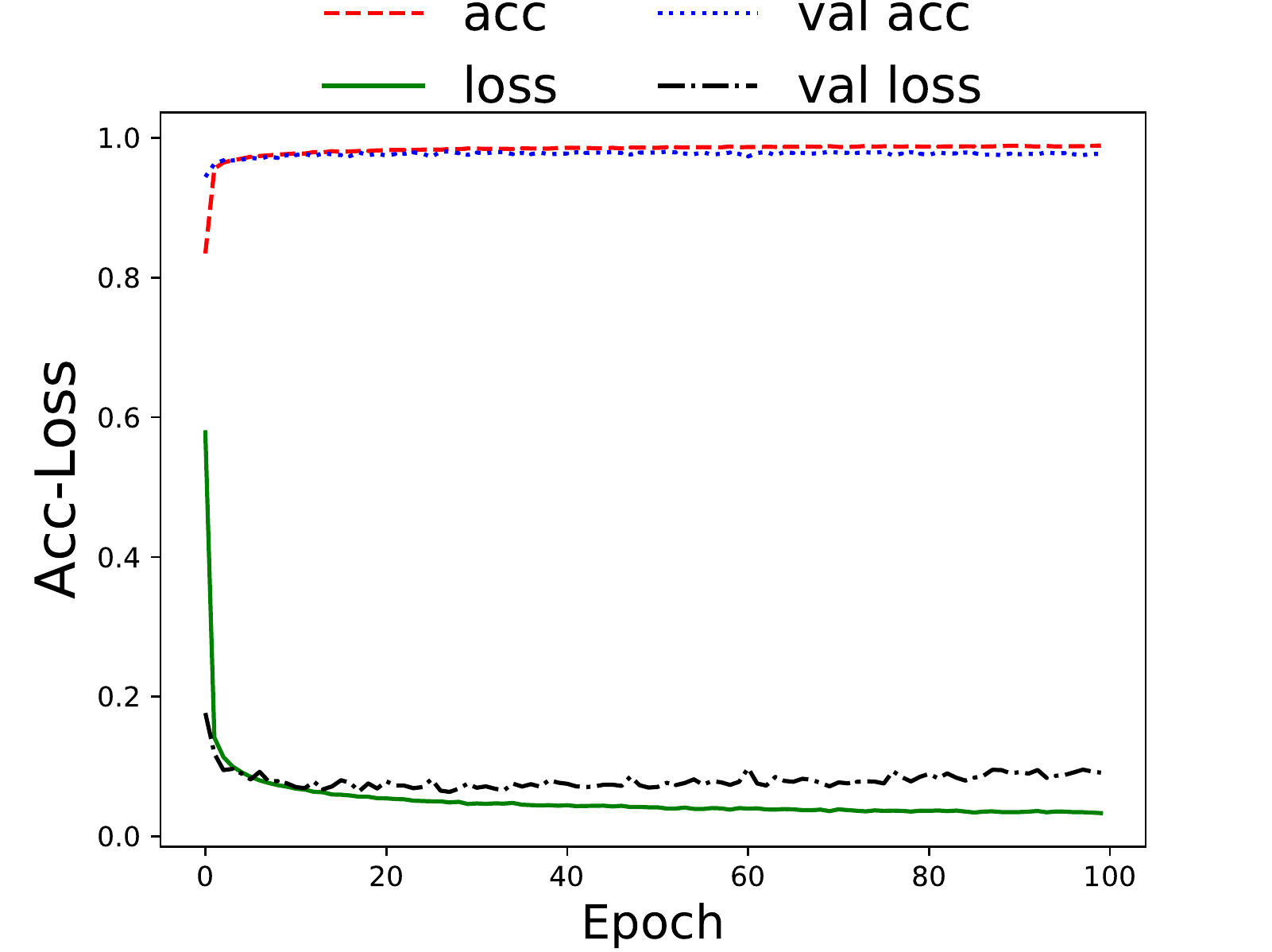}}
   \subfigure[LeNet-4]{
   \label{figure6b}
   \includegraphics[width=0.32\linewidth]{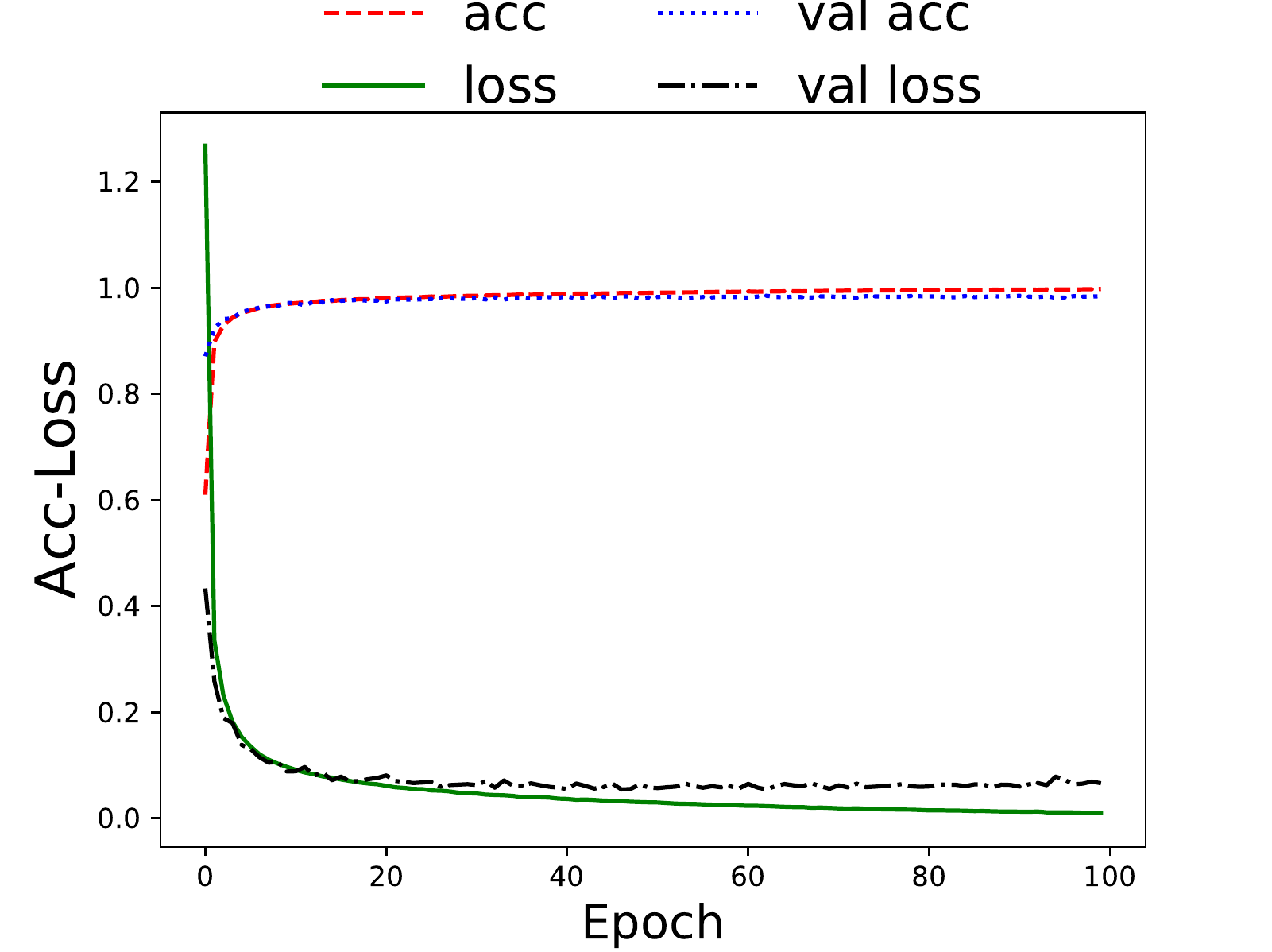}}
   \subfigure[LeNet-5]{
   \label{figure6c}
   \includegraphics[width=0.32\linewidth]{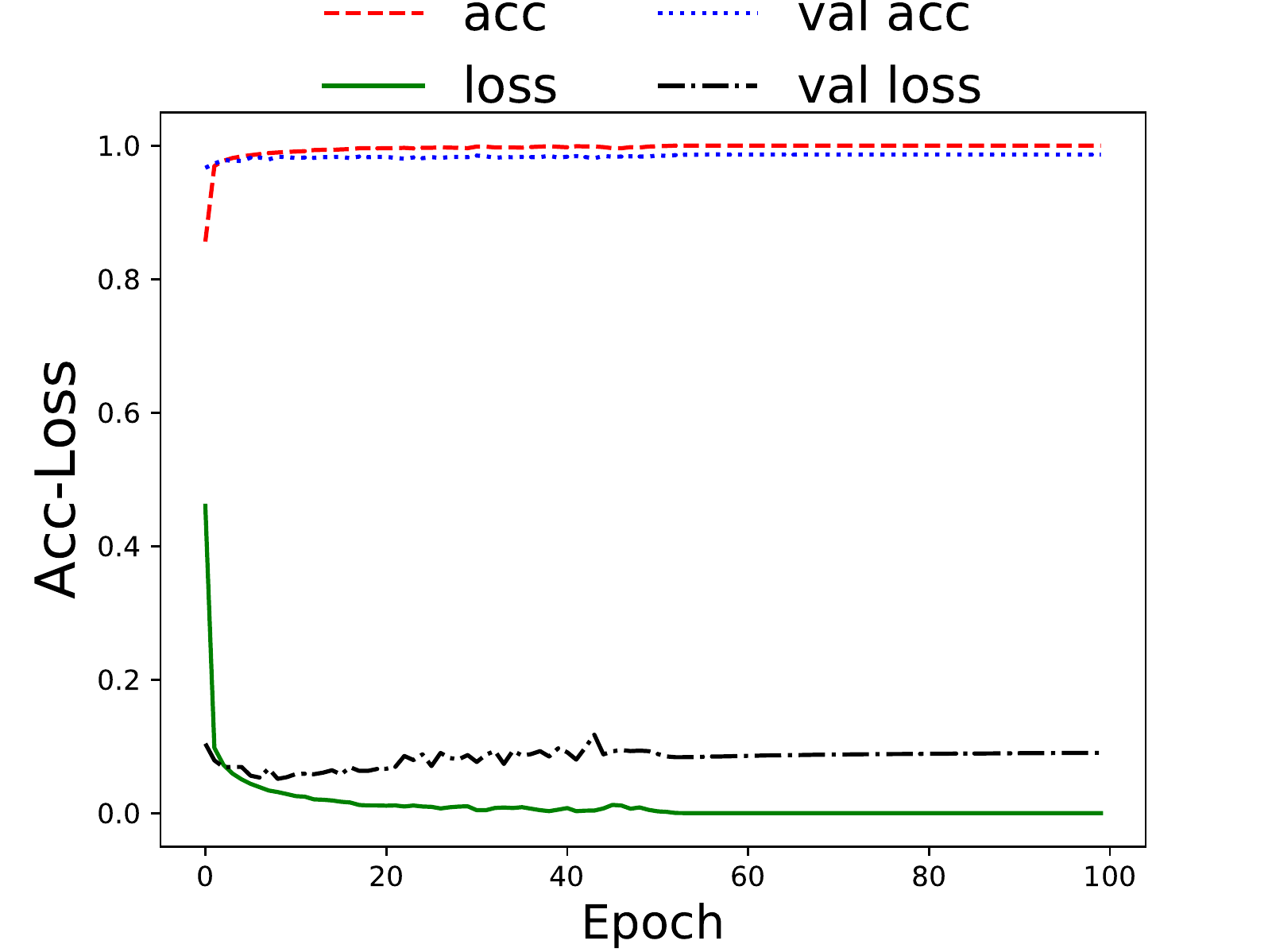}}
   \subfigure[VGG-16]{
   \label{figure6d}
   \includegraphics[width=0.32\linewidth]{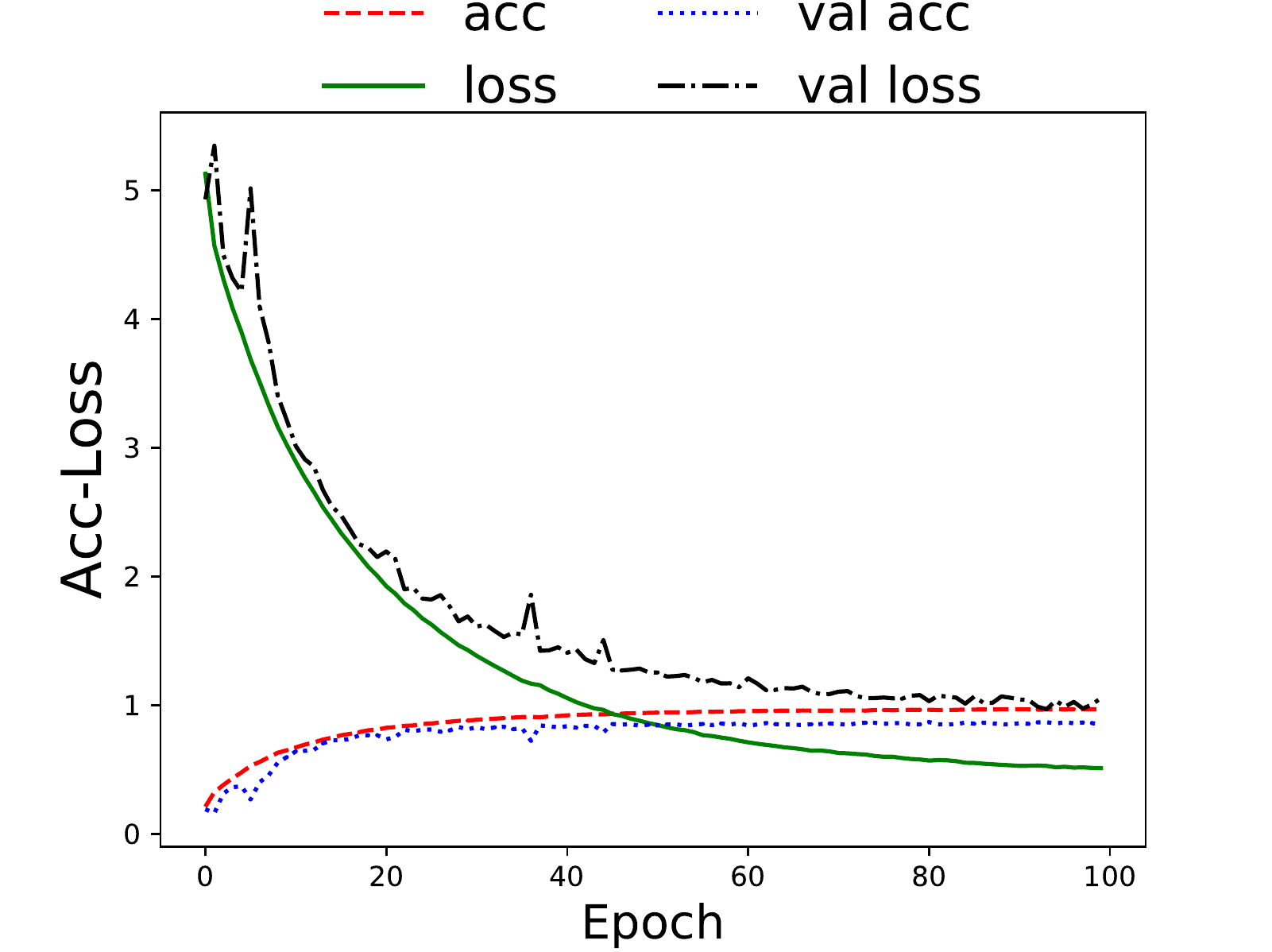}}
   \subfigure[VGG-19]{
   \label{figure6e}
   \includegraphics[width=0.32\linewidth]{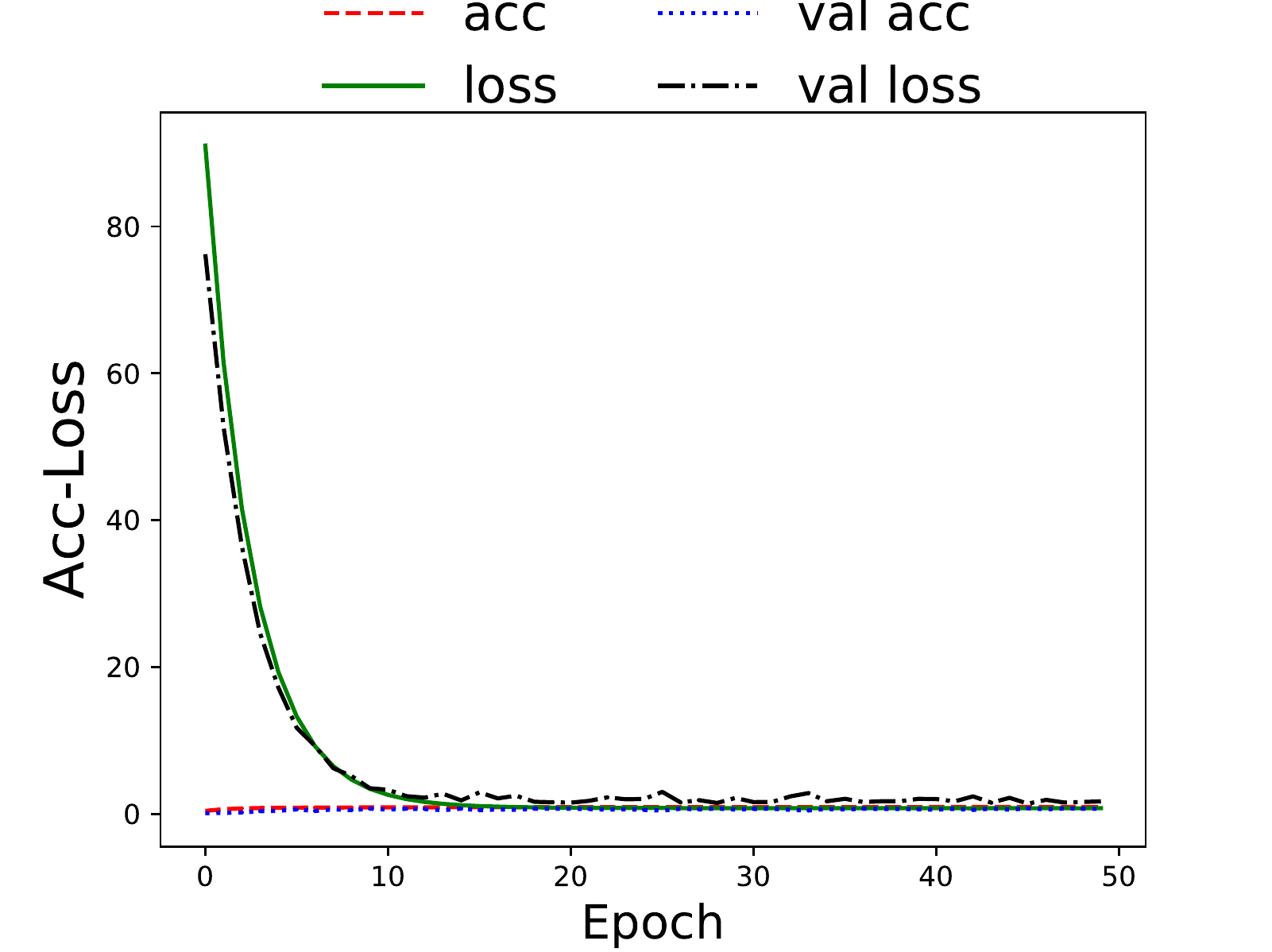}}
   \subfigure[ResNet-20]{
   \label{figure6f}
   \includegraphics[width=0.32\linewidth]{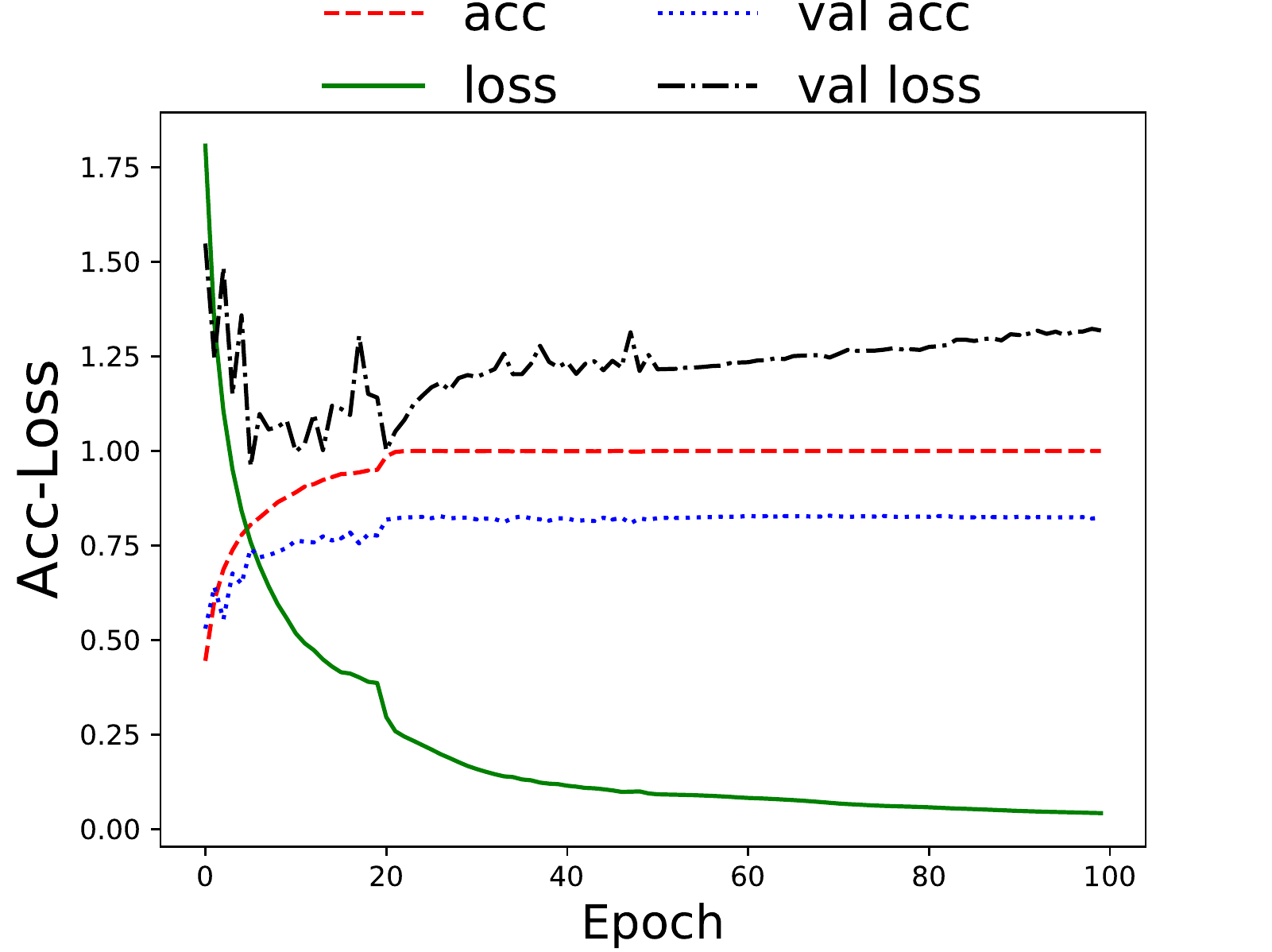}}
   \caption{Model training record diagram:(a) LeNet-1, training on MINIST Data Set, when epoch=100, (b) LeNet-4, training on MINIST Data Set, when epoch=100, (c)LeNet-5, training on MINIST Data Set, when epoch=100, (d)VGG-16, training on CIFAR-10 Data Set, when epoch=100, (e)VGG-19, training on CIFAR-10 Data Set, when epoch=50, (f)ResNet-20, training on CIFAR-10 Data Set, when epoch=100.} \label{train_loss}
\end{figure*}

In Fig.~\ref{train_loss}, we show the training loss, training accuracy, validation loss, and validation accuracy of each model. As can be seen in the figure, during the training process of LeNet-5, with the increase of training times, the loss value of the model gradually decreases, and the accuracy is getting higher and higher. This shows that with the increase of training time, the model can fit the data well, and the model can accurately classify the dataset. We follow the criterion of machine learning, and then choose the competitive DNN models as the research object for the fuzzy test under the condition of fitting.

\subsubsection{Generalization Ability}
To answer \emph{RQ1}, we compare \emph{CAGFuzz} with the existing model-based approach \emph{FGSM}. In the experiment, we enhanced \emph{FGSM} by adding the coverage feedback to the generated adversarial examples. In this way, \emph{FGSM} has the same coverage-guided test approach as \emph{CAGFuzz}.
We choose MNIST and CIFAR-10 data sets as the sampling set. For MNIST, we sample 10 examples for each class in the training set and 4 examples for each class in the test set. Since the DNN models used to classify CIFAR-10 data set have a large scale of weight parameters, 10 training examples are not enough to achieve the training effect. Therefore, for CIFAR-10, we sample 100 examples for each class in the training set and 10 examples for each class in the test set. 

Based on the LeNet-1 model, we use \emph{FGSM} to generate an adversarial example for each of the 10 examples in the training set, and also use \emph{AEG} to generate an adversarial example for each training example. First, the original data set is used to train and test the LeNet-1 model. We set the epoch as 50 and the learning rate as 0.05. Then, the adversarial examples generated by \emph{CAGFuzz} and \emph{FGSM} are added to the training set to retrain LeNet-1 with the same parameters. Finally, the above two steps are repeated, but the model is replaced by LeNet-4 or LeNet-5.

Similar to generating adversarial examples based on LeNet-1, we perform the same experiment on LeNet-4 and LeNet-5. Because of the uncertainty during the model training process, we train the model repeatedly 5 times in the same setting, and take the average of these results as the final accuracy of our experiments.
For example, the 5 times accuracy of the ResNet-20 model under FGSM-R20 fluctuates; therefore, we take the average value.
Table~\ref{gb} shows the accuracy of the three models on the original data set, FGSM-Le1, FGSM-Le4, FGSM-Le5 and CAGFuzz-dataset. Among them, ``FGSM-Le1" refers to the data set generated by \emph{FGSM} method. "CAGFuzz-dataset" refers to the data set generated based on \emph{CAGFuzz}.
From the table, it can be seen that the accuracy of the adversarial examples generated by \emph{FGSM} based on a specific model is improved higher than that of other models. For example, after retraining LeNet-1 based on FGSM-Le1, the accuracy is 70.6\%. After retraining LeNet-1 based on FGSM-Le4 and FGSM-Le5, the accuracy is 66.6\% and 68.6\%. Analyzing all data in the Table~\ref{gb}, we can see that after retraining three models based on CAGFuzz-dataset, the accuracy of the models are all high, namely, 72.6\%, 72\% and 74.3\%.
In the same way, similar results are obtained when applied to the CIFAR10 data set, and the final results are shown in Table~\ref{gb_cifar10}. We can see that after retraining three models based on CAGFuzz-dataset, the accuracy of the model is mostly higher than the maximum accuracy of the retraining model based on \emph{FGSM}. In the ResNet-20 model, the final accuracy of \emph{CAGFuzz} retraining model is 39.2\%, which is a little worse than FGSM-R20, but much better than FGSM-V16 and FGSM-V19.

\emph{$\rhd$ Answer to \textbf{RQ1}:} Taking the MNIST and CIFAR-10 data sets as examples, we prove
that the adversarial examples generated based on target model (such as \emph{FGSM}) only improve the accuracy of this special model better, and the improvement on other models is limited. On the contrary, \emph{CAGFuzz} can generate adversarial examples based on data, and this can improve the accuracy of all the models to almost the same degree. Summarizing, the adversarial examples generated based on \emph{CAGFuzz} has better generation ability of the adversarial examples generated based on target model.


\begin{table}[ht]
\caption{The accuracy of the three models on the MNIST data set, adversarial examples generated based on \emph{FGSM} and adversarial examples generated based on \emph{CAGFuzz}(\%)}\label{gb}
\centering
\renewcommand\arraystretch{1.5}
\begin{tabular}{p{1cm}<{\centering} p{1cm}<{\centering} p{1cm}<{\centering} p{1cm}<{\centering} p{1cm}<{\centering} p{1cm}<{\centering}}
  \hline
  Model & Orig. dataset & FGSM-Le1 & FGSM-Le4 & FGSM-Le5 & CAGFuzz-dataset\\
  \hline
  LeNet1 & 59   & \textbf{70.6} &66.6 &68.6 & \textbf{72.6}\\
  LeNet4 & 62.6 & 66.6 &\textbf{71.6} &68.2 & \textbf{72}\\
  LeNet5 & 60.6 & 69.3 &64.6 &\textbf{71}   & \textbf{74.3}\\
  \hline
\end{tabular}
\end{table}

\begin{table}[ht]
\caption{The accuracy of the three models on the CIFAR10 data set, adversarial examples generated based on \emph{FGSM} and adversarial examples generated based on \emph{CAGFuzz}(\%)}\label{gb_cifar10}
\centering
\renewcommand\arraystretch{1.5}
\begin{tabular}{p{1cm}<{\centering} p{1cm}<{\centering} p{1cm}<{\centering} p{1cm}<{\centering} p{1cm}<{\centering} p{1cm}<{\centering}}
  \hline
  Model & Orig. dataset & FGSM-V16 & FGSM-V19 & FGSM-R20 & CAGFuzz-dataset\\
  \hline
  VGG16 & 19 & \textbf{28.2} &21.8 &24 & \textbf{30.2}\\
  VGG19 & 10 & 18.4 &\textbf{25.6} &21.4 & \textbf{27}\\
  ResNet20 & 15 & 33.8 &36.8 &\textbf{40} & \textbf{39.2}\\
  \hline
\end{tabular}
\end{table}


\subsubsection{Neuron Coverage}\label{subsec_NCResults}
To answer \emph{RQ2}, we use the training data set of each model as the input set to calculate the original neural coverage, and the generated adversarial example set as the input set to calculate the neural coverage of \emph{CAGFuzz}.

Obviously, the adversarial examples generated by \emph{AEG} can effectively improve the neuron coverage of the target DNNs. To further validate the effectiveness of \emph{CAGFuzz} in improving neuron coverage, we also compare it with other three approaches. 
Table~\ref{Tab3} lists the original neuron coverage of each model and the neuron coverage using the different approaches.
It can be seen from the table that in the MNIST data set, the \emph{FGSM} approach does not improve the coverage of the model. For the LeNet-1 and LeNet-4 models, the coverage improvement of \emph{CAGFuzz} is not better than \emph{DeepHunter} and \emph{DeepXplore}. However, the coverage improvement effect of \emph{CAGFuzz} in the LeNet-5 model is obvious better than the other two approaches.
In the CIFAR-10 data set, the coverage improvement of the \emph{FGSM} approach is also not good, and even worse than the coverage of the original examples. The coverage improvement of \emph{CAGFuzz} is generally better than other approaches, in addition to the ResNet-20 model, \emph{DeepHunter} increases the coverage to 78.62\%, while \emph{CAGFuzz} only increases to 75.74\%.
In the ImageNet data set, \emph{CAGFuzz} can improve the model coverage better than all other approaches.

\emph{$\rhd$ Answer to \textbf{RQ2}:} In conclusion, \emph{CAGFuzz} can effectively generate adversarial examples and these adversarial examples can improve neuron coverage for the target DNN. Due to the deep feature constraint, the adversarial examples generated by \emph{CAGFuzz} can significantly improve the  neuron coverage in the model with deep depth and large number of neurons.

\begin{table}[ht]
\caption{Comparison of \emph{CAGFuzz}, \emph{FGSM}~\cite{goodfellow2014explaining}, \emph{DeepHunter}~\cite{Xie2018DeepHunter} and \emph{DeepXplore}~\cite{Pei2017DeepXplore} in Increasing the Neuron Coverage of Target DNNs.}\label{Tab3}
\centering
\renewcommand\arraystretch{1.5}
\begin{tabular}{p{1cm}<{\centering} p{1cm}<{\centering} p{1cm}<{\centering} p{1.1cm}<{\centering} p{1.1cm}<{\centering} p{1.1cm}<{\centering}}
  \hline
  DNN Model & Orig. NC(\%) & \emph{FGSM} NC(\%) & \emph{DeepHunter} NC(\%) & \emph{DeepXplore} NC(\%) & \emph{CAGFuzz} NC(\%)\\
  \hline
  LeNet1      & 38.46  & 38.46  &53.84  &57.69  & \textbf{46.15}\\
  LeNet4      & 72.41  & 72.41  &80.17  &81.89  & \textbf{79.31}\\
  LeNet5      & 86.44  & 86.44  &88.98  &87.28  & \textbf{93.22}\\
  \hline
  VGG16    & 50.99  & 47.30  &59.71  &55.39  & \textbf{62.32}\\
  VGG19    & 55.33  & 55.47  &57.34  &56.02  & \textbf{58.51}\\
  ResNet20 & 75.04  & 75.33  &78.62  &75.37  & \textbf{75.74}\\
  \hline
  VGG16    & 13.33  & 13.91  &14.07  &13.68  & \textbf{14.54}\\
  VGG19    & 13.98  & 14.74  &15.24  &14.01  & \textbf{16.36}\\
  ResNet50 & 76.88  & 77.28  &76.44  &77.96  & \textbf{78.26}\\
  \hline
\end{tabular}
\end{table}

\subsubsection{Error Behavior Discovery} \label{subsec_ErrorResults}
To answer \emph{RQ3}, we sample correctly classified examples by DNN models from the test set of each dataset. Based on these correctly classified examples, we generated adversarial examples for each example through the \emph{AEG} of each dataset. The examples we selected are all positive examples with correct classification. We can confirm that all the generated adversarial examples should also be classified correctly, because the deep semantics information of the adversarial examples and the original examples are consistent. The ``positive examples'' generated by \emph{AEG} are input into the corresponding classifier model for classification. If there are errors or classification errors, a potential defect of the classification model can be found. 
We define the original correct example as $Image_{orig}$ and the corresponding adversarial example as $Image_{adv}=\{Image_{1}, Image_{2},...,Image_{10}\}$. The original example $Image_{orig}$ is classified correctly in the target DNN model, consequently $Image_{i}$ should also be classified correctly, where $Image_{i}\in Image_{adv}$. If the $Image_{i}$ classification of an adversarial example is wrong, we consider this to be an error behavior of the target DNN.

We choose a quantitative measure to evaluate the effectiveness of \emph{CAGFuzz} in detecting erroneous behaviors in different models. As mentioned above, we take 2000 examples, which are verified to be correct from each data set. Then we use the four approaches mentioned to mutation these examples, and generate 2000 adversarial examples for our experiments. Table~\ref{Tab4} shows the number of erroneous behaviors found by different datasets under the guidance of neuron coverage. In addition, we also list the number of errors found by \emph{FGSM}~\cite{goodfellow2014explaining}, \emph{DeepHunter}~\cite{Xie2018DeepHunter}, and \emph{DeepXplore}~\cite{Pei2017DeepXplore} in each data set.

\begin{table}[!h]
\caption{Number of erroneous behaviors reported by \emph{FGSM}~\cite{goodfellow2014explaining}, \emph{DeepHunter}~\cite{Xie2018DeepHunter}, \emph{DeepXplore}~\cite{Pei2017DeepXplore}, and \emph{CAGFuzz} across 2000 adversarial examples.}\label{Tab4}
\centering
\renewcommand\arraystretch{1.5}
\begin{tabular}{p{1.5cm}<{\centering} p{1cm}<{\centering} p{1.3cm}<{\centering} p{1.3cm}<{\centering} p{1.3cm}<{\centering}}
  \hline
  {Data Sets} & \emph{FGSM} & \emph{DeepHunter} &\emph{DeepXplore} &\emph{CAGFuzz} \\
  \hline
  MNIST      & 162  & 670   &34  &\textbf{894}\\
  CIFAR-10   & 69   & 193   &20  &\textbf{284}\\
  ImageNet   & 278  & 456   &18  &\textbf{720}\\
  \hline
  SUM        & 509  & 1319  &72  &\textbf{1898} \\
  \hline
\end{tabular}
\end{table}

As can be seen from Table~\ref{Tab4}, the \emph{DeepXplore's} ability to detect potential errors is poor, and its performance in each data set is not ideal. The total number of potential errors found in the three data sets is 72. Compared with the other three approaches, \emph{CAGFuzz} has a stronger ability on finding potential errors in the model. It has good results in all the three datasets, and a total of 1898 potential errors in the model have been found.

\emph{$\rhd$ Answer to \textbf{RQ3}:}  With neuron coverage guided adversarial examples, and based on the same model and the same positive examples, \emph{CAGFuzz} can find more potential errors in the model.


\begin{figure}[t]
 \centering
 \includegraphics[width=0.8\linewidth]{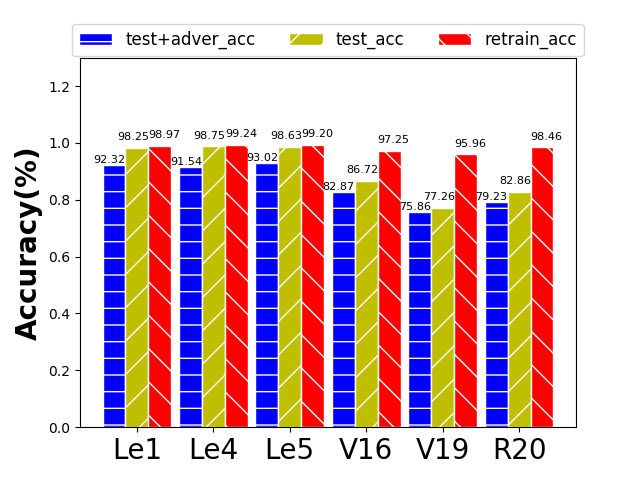}\\
 \caption{Improvement of accuracy and robustness after retraining model.}\label{acc_increase}
\end{figure}

\begin{figure*}[ht]
\centering
   \subfigure[LeNet-1]{
   \label{figure9a}
   \includegraphics[width=0.32\linewidth]{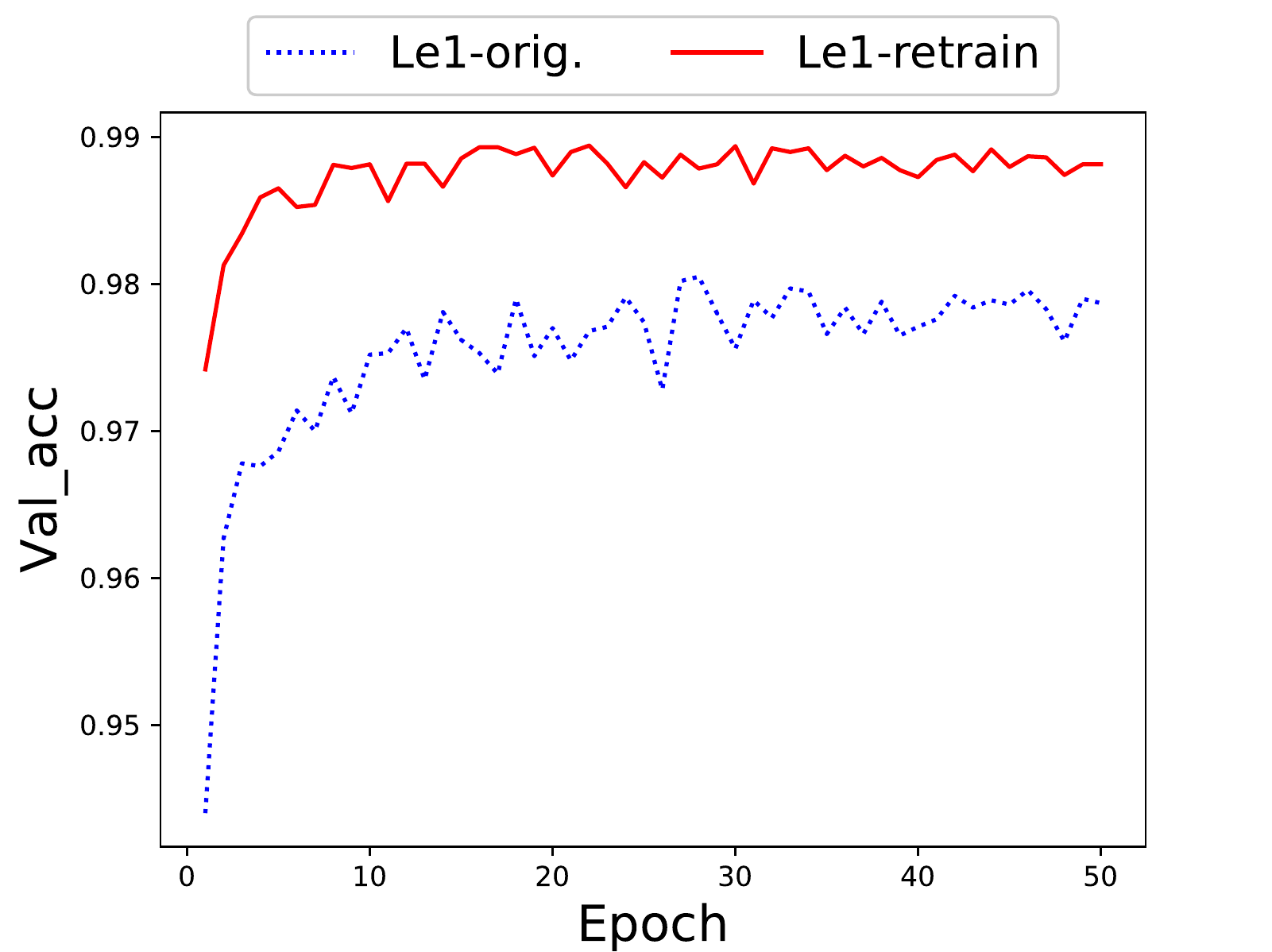}}
   \subfigure[LeNet-4]{
   \label{figure9b}
   \includegraphics[width=0.32\linewidth]{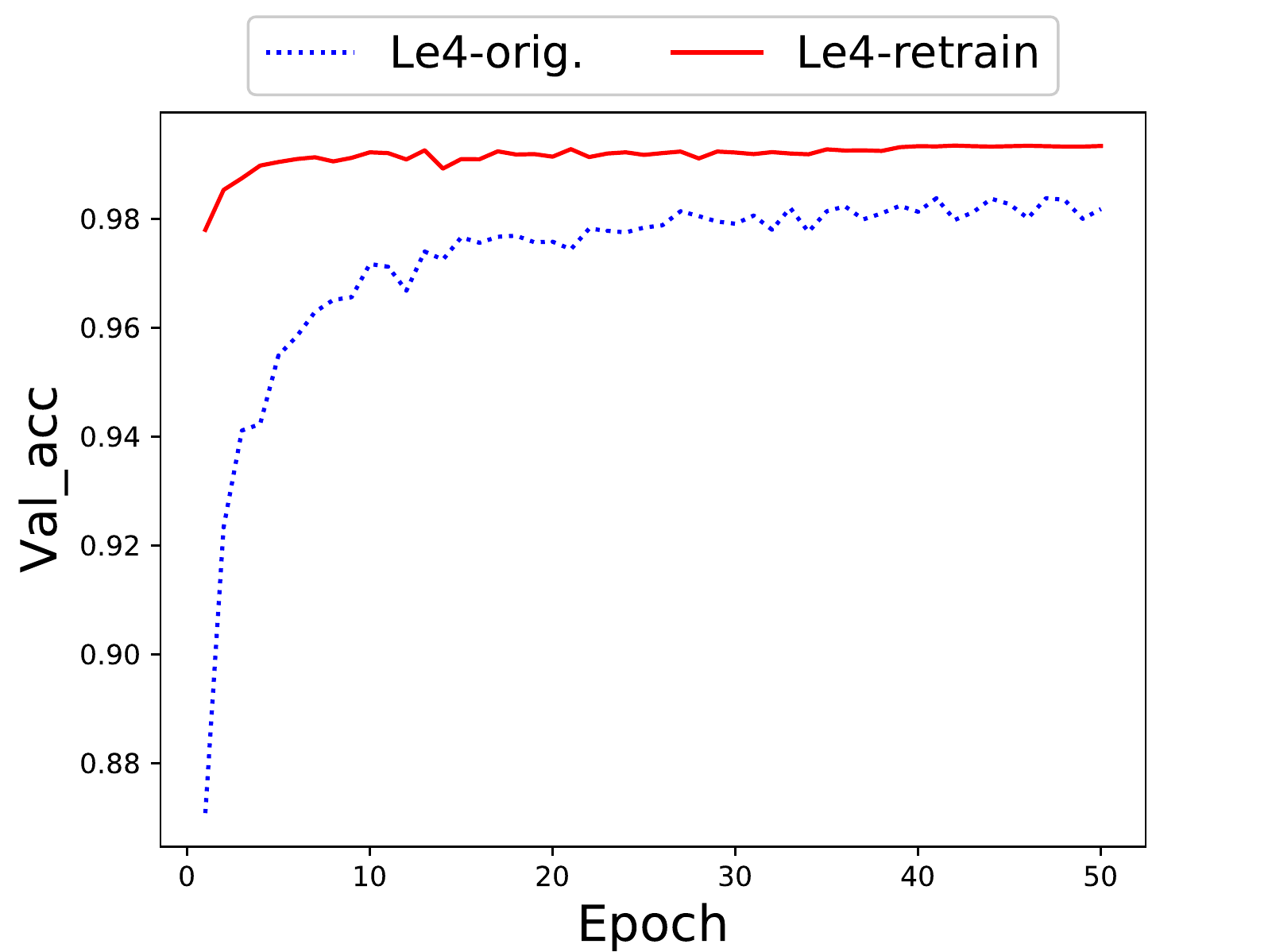}}
   \subfigure[LeNet-5]{
   \label{figure9c}
   \includegraphics[width=0.32\linewidth]{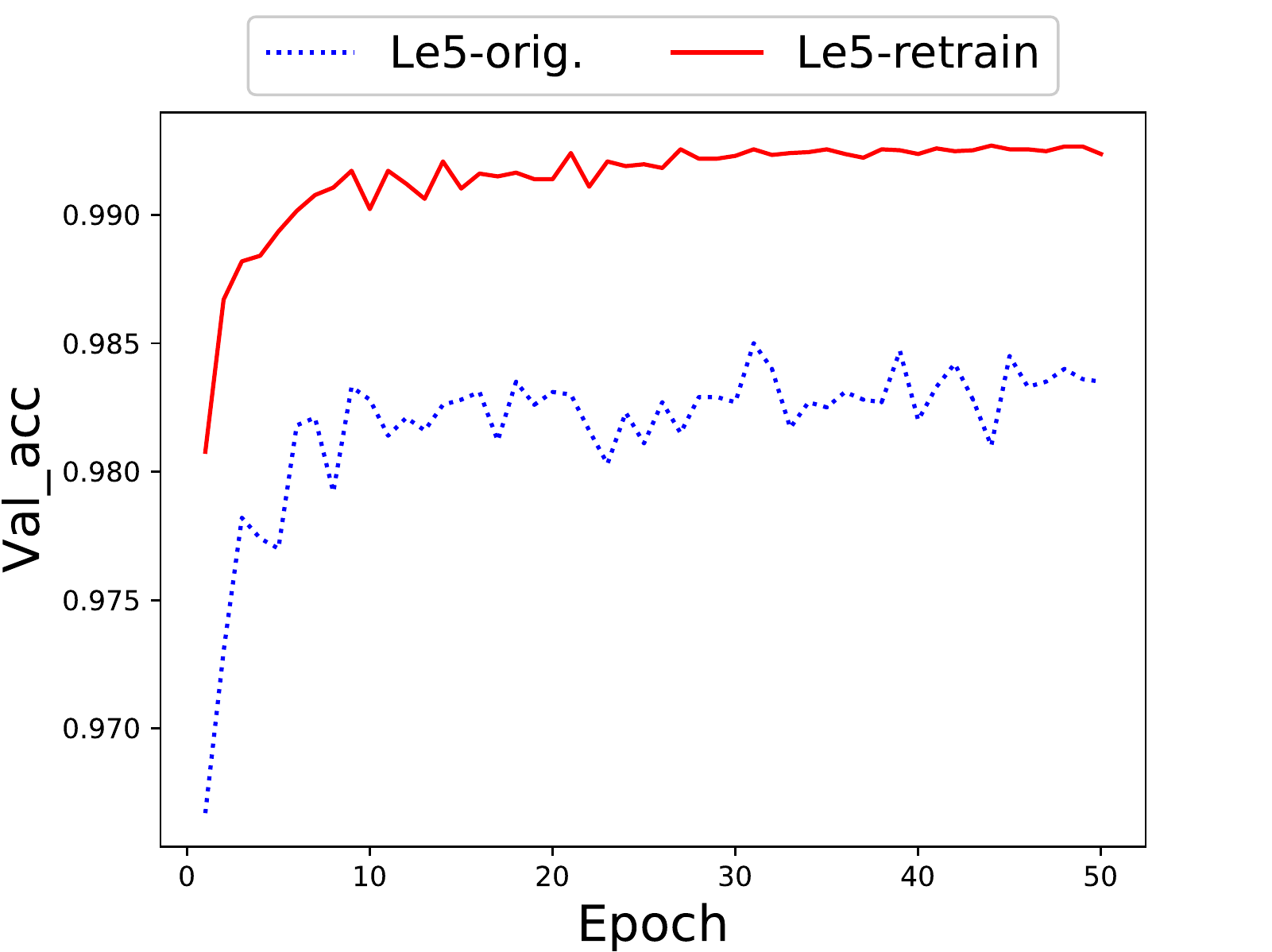}}
   \subfigure[VGG-16]{
   \label{figure9d}
   \includegraphics[width=0.32\linewidth]{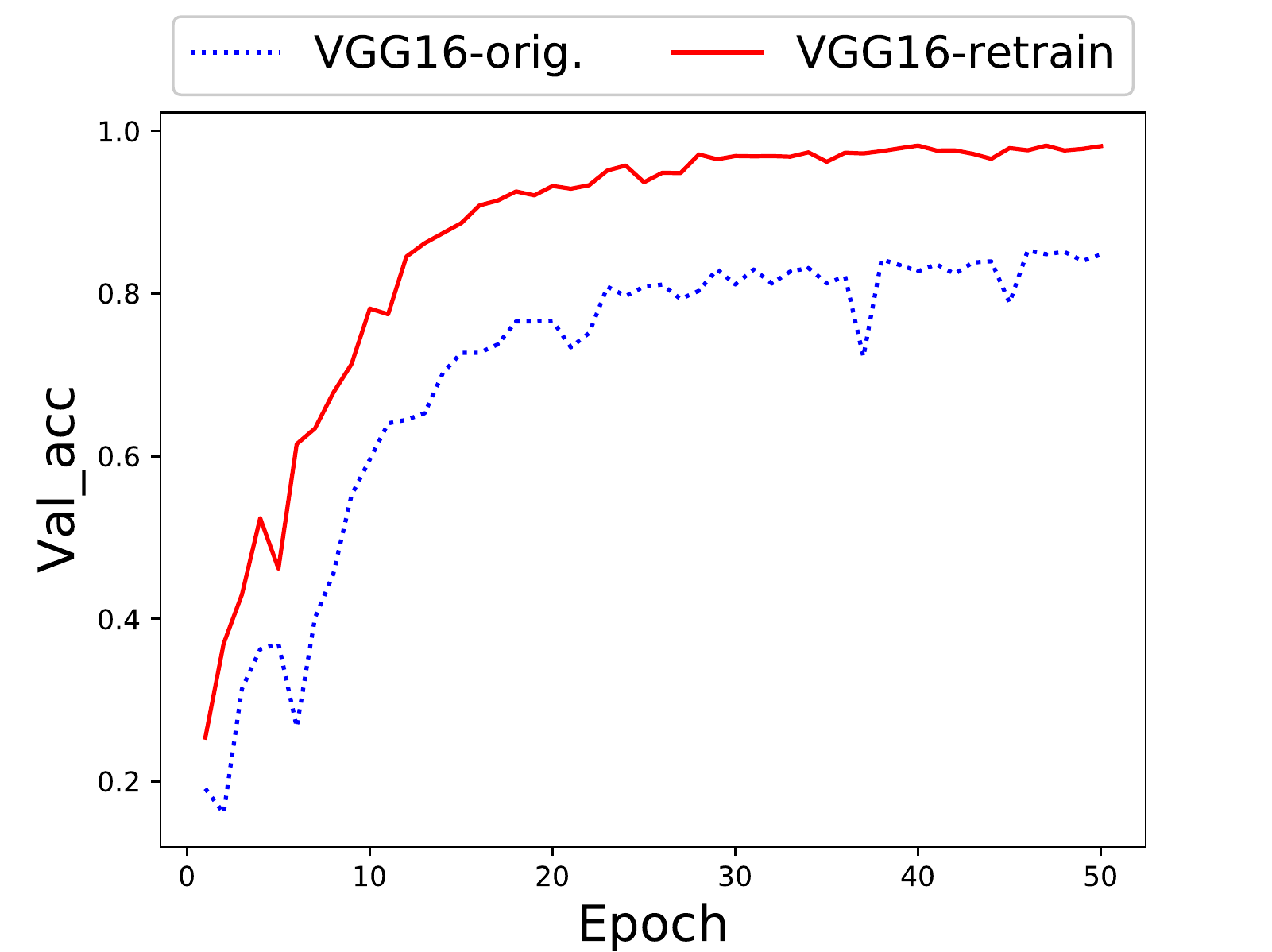}}
   \subfigure[VGG-19]{
   \label{figure9e}
   \includegraphics[width=0.32\linewidth]{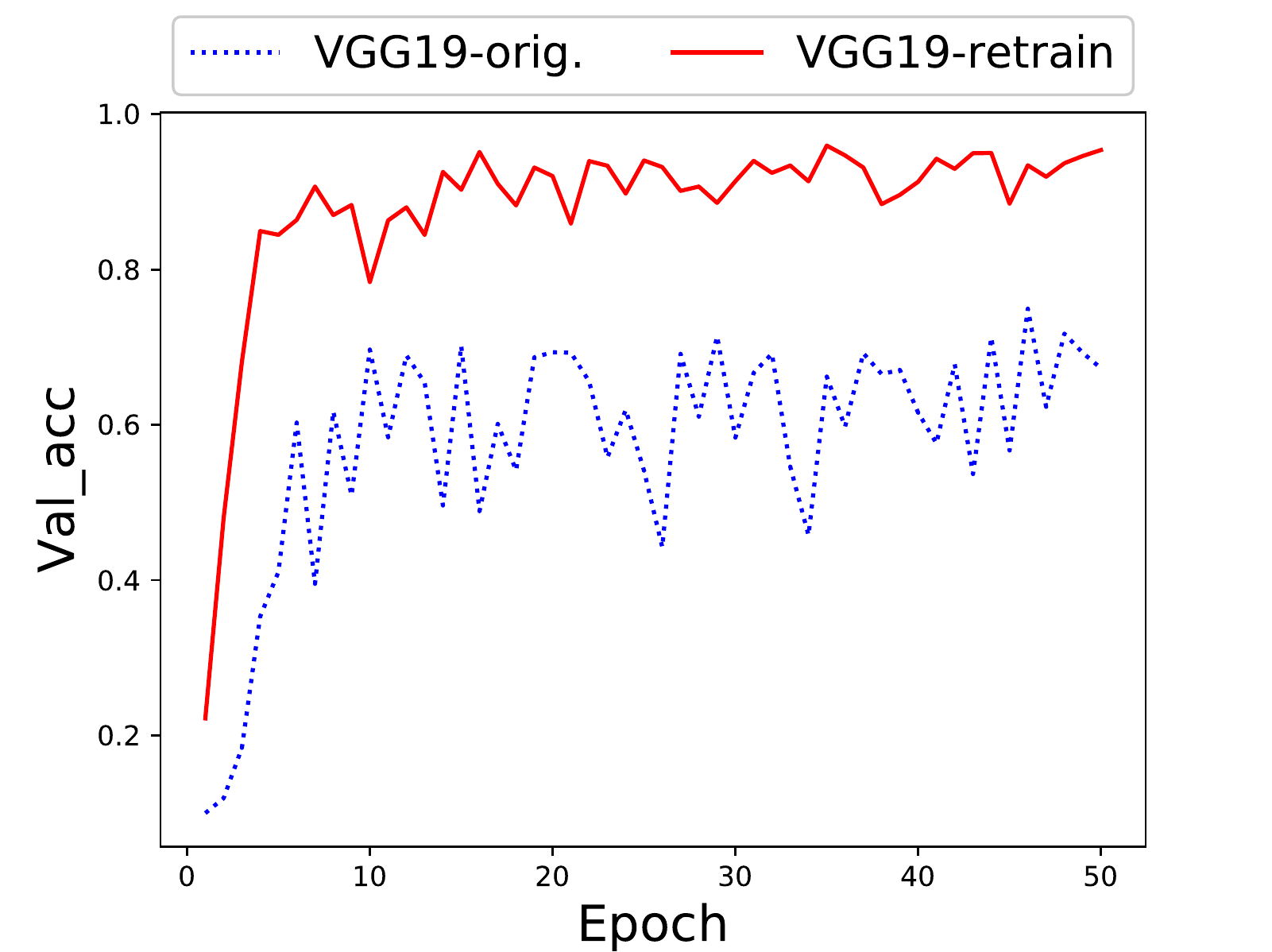}}
   \subfigure[ResNet-20]{
   \label{figure9f}
   \includegraphics[width=0.32\linewidth]{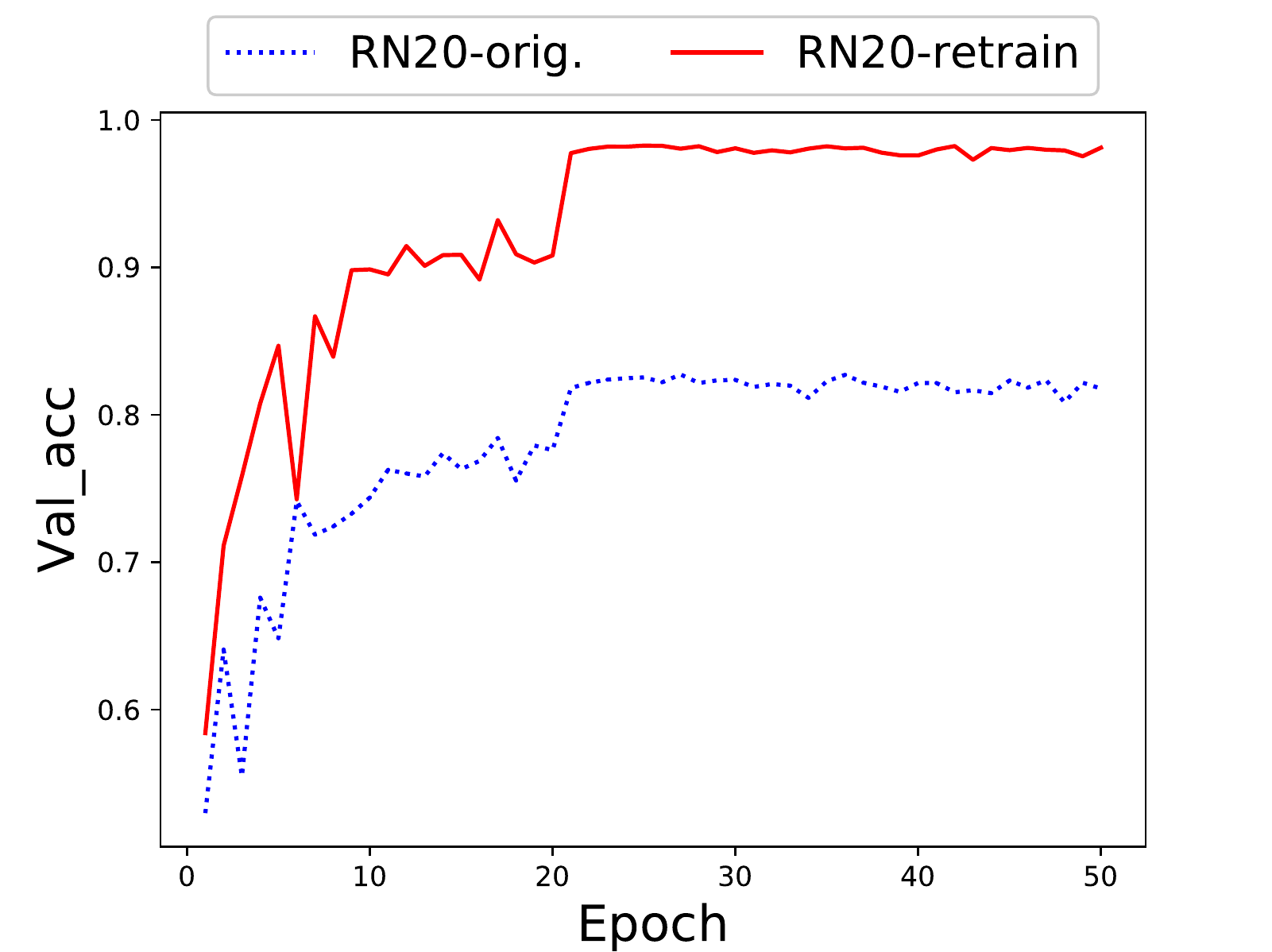}}
   \caption{Validation Set Accuracy Contrast Diagram of Each Model in the Training Process: (a) LeNet-1, training on MINIST Data Set, when epoch=50, (b) LeNet-4, training on MINIST Data Set, when epoch=50, (c) LeNet-5, training on MINIST Data Set, when epoch=50, (d) VGG-16, training on CIFAR-10 Data Set, when epoch=50, (e)VGG-19, training on CIFAR-10 Data Set, when epoch=50, (f)ResNet-20, training on CIFAR-10 Data Set, when epoch=70.} \label{acc_contrast}
\end{figure*}



\subsubsection{Accuracy and Robustness} \label{subsec_AccResults}
To answer \emph{RQ4}, we add adversarial examples generated by \emph{CAGFuzz} to the training set to retrain the DNN model and measure whether it can improve the accuracy of the target DNN.
We select the MNIST and CIAR-10 data sets as our experimental data sets, and we select three DNN models, LeNet-1, 4, 5,  and the VGG-16, VGG-19, and ResNet-20 models as experimental models. We retrain the DNN model by mixing 65\% of the adversarial example set and the original training set, and then validate the DNN model with the remaining 35\% of the adversarial example set and the original test set on the original model and the retraining model. Because of the limitation of the size of the picture, in Fig.~\ref{acc_increase}, we abbreviate the model name. For example, the model LeNet-1 is abbreviated to Le1, the model VGG-16 is abbreviated to V16, and the model ResNet20 is abbreviated to R20. In Fig.~\ref{acc_increase}, ``test\_acc" represents the accuracy of the model on the original test set, ``test+adver\_acc" represents the accuracy of the model on the test set with adversarial examples (the model is still the original one), and ``retrain\_acc" represents the accuracy of the model after retraining the model with the adversarial examples. It can be seen from the comparison of ``test\_acc" and ``test+adver\_acc" that the robustness of the original model is very poor. After adversarial examples are added into the test set, the accuracy of the model decreases obviously. For example, the accuracy of the LeNet-5 model decreases from 98.63\% to 93.02\% and from 5.69\% on the original basis.
In the VGG-19 model, the accuracy of the model decreases from 77.26\% to 75.86\%. The comparison of ``test\_acc" and ``retrain\_acc" shows that the accuracy of the models has been greatly improved after retraining the model with the adversarial examples, especially for the VGG model with deeper layers. For example, from Fig.~\ref{acc_increase}, we can see that the accuracy of the VGG-19 network has increased from 77.26\% to 95.96\%, with an increase of 24.2\%. In general, we can see that \emph{CAGFuzz} can not only improve the robustness of the model, but also improve the accuracy of the model, especially for the model with deeper network layer.


In the experiments, we further analyze the accuracy of the retraining model and the original model during the training process, and evaluate the validity of the adversarial examples generated by \emph{CAGFuzz} from the change of the validation accuracy. Fig.~\ref{acc_contrast} shows the changes of validation accuracy of different models during training. The original structure parameters and learning rate of each model are kept unchanged, and the new data set we reconstituted is used for retraining. During the training process, the validation accuracy and the original validation accuracy of the same epoch are compared. It can be found that under the same epoch, the validation accuracy of the retraining model is higher than that of the original model, and the convergence speed of the retraining model is faster. Moreover, it can also be found from the figure that the retraining model is more stable and has a smaller change range during the training process.

In addition, we can see that the trend of the retrained model is basically consistent with the original model, which shows that the accuracy of the model can be greatly improved without affecting the internal structure and logic of the model. For example, in Fig.~\ref{figure9d}, the accuracy of the original model drops suddenly when epoch = 6, and the retraining model also continues this change. In Fig.~\ref{figure9f}, the original model presents a three-stage upgrade, which is reflected in the retraining model at the same time.

\begin{figure*}[ht]
 \centering
 \subfigure[LeNet-1]{
   \label{figurea}
   \includegraphics[width=0.32\linewidth]{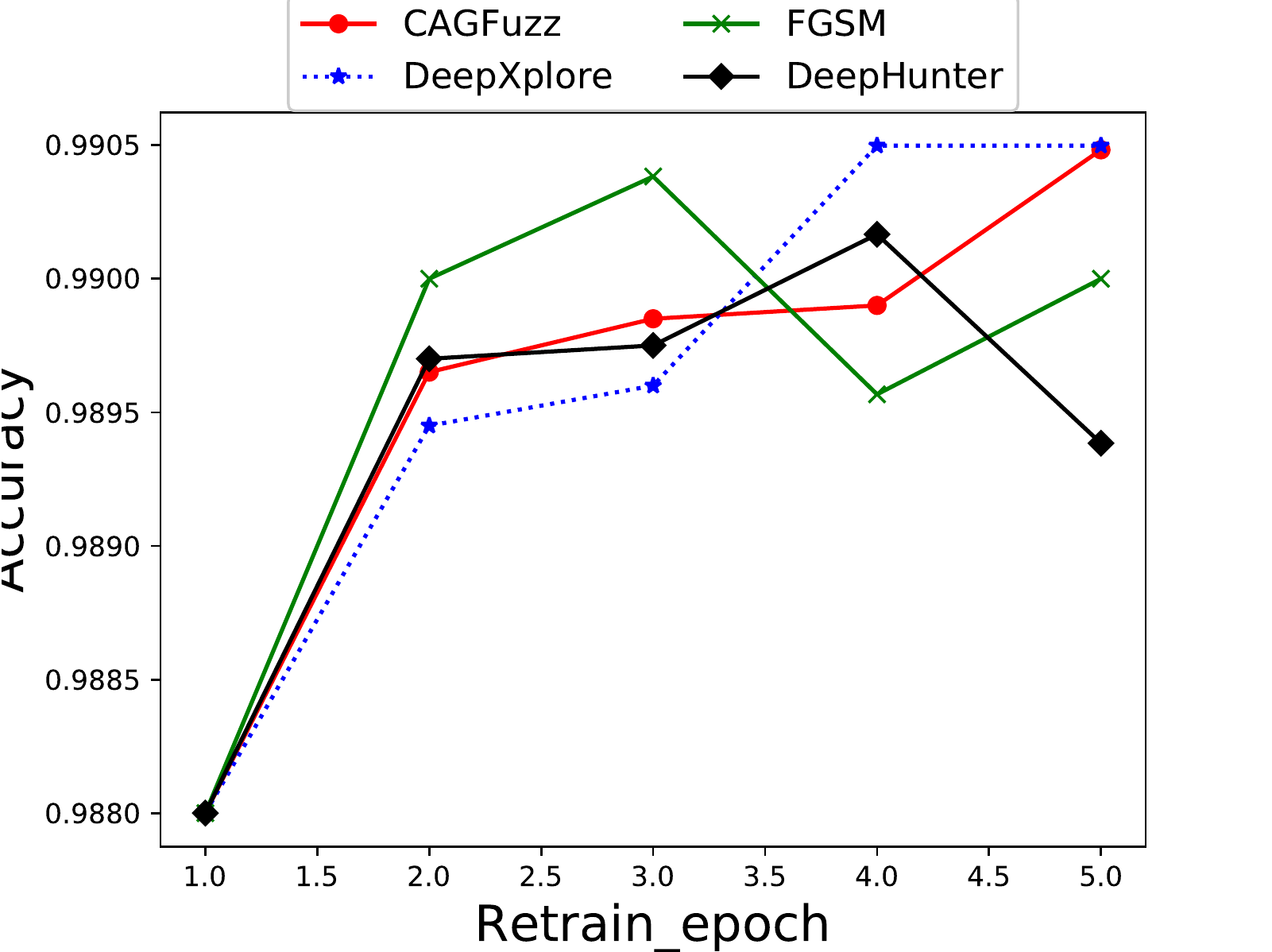}}
   \subfigure[LeNet-4]{
   \label{figureb}
   \includegraphics[width=0.32\linewidth]{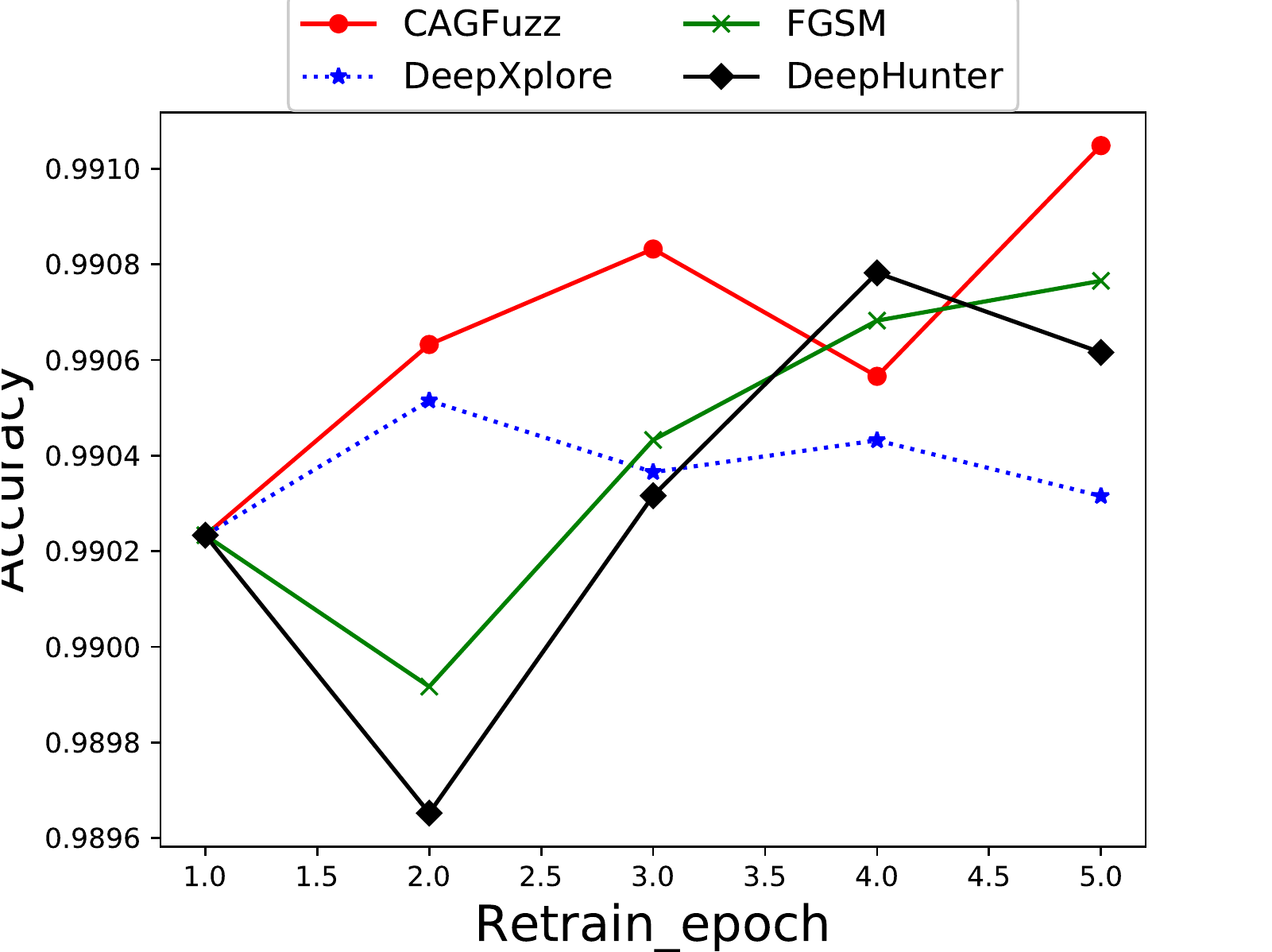}}
   \subfigure[LeNet-5]{
   \label{figurec}
   \includegraphics[width=0.32\linewidth]{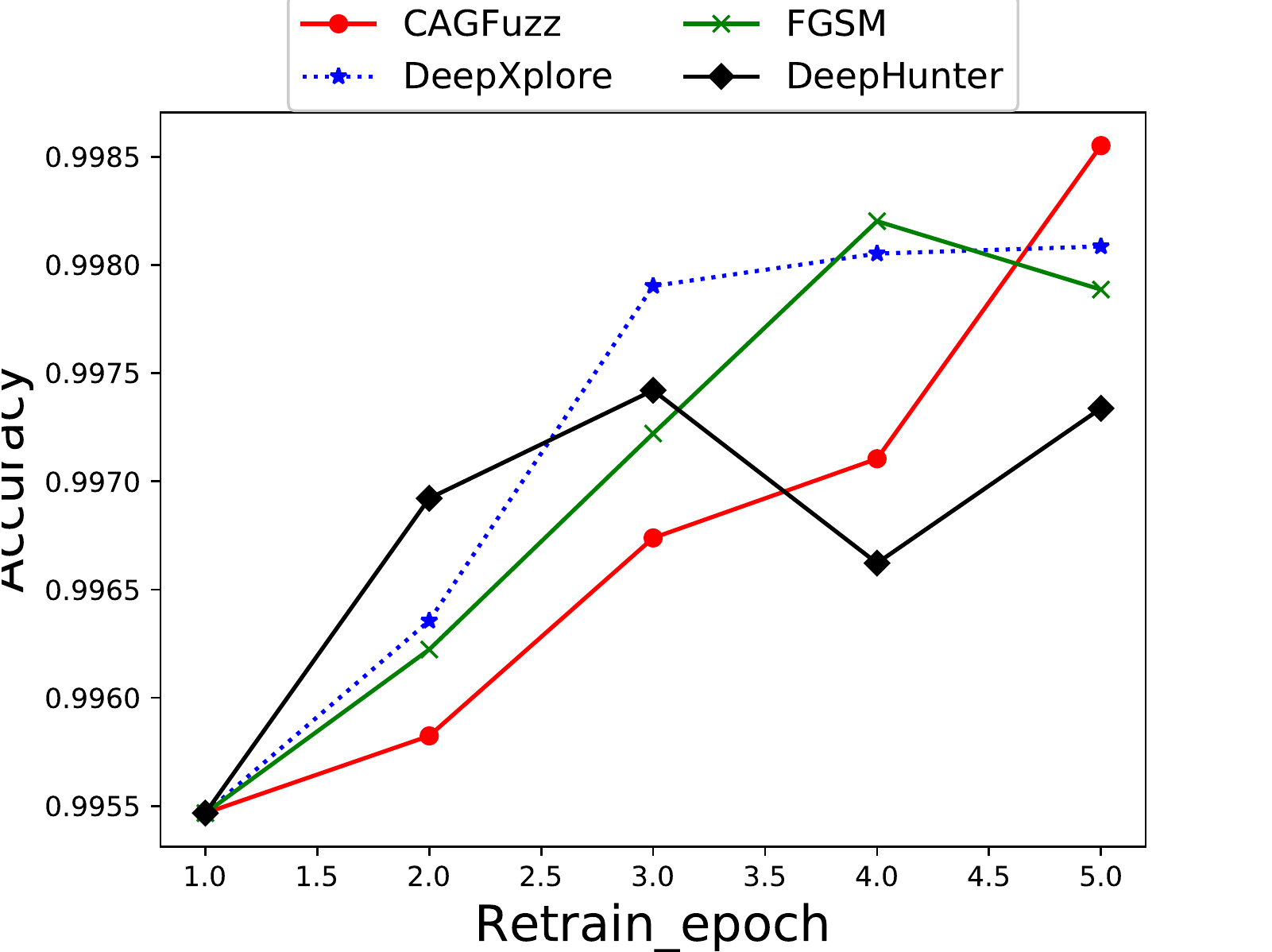}}
   \subfigure[VGG-16]{
   \label{figured}
   \includegraphics[width=0.32\linewidth]{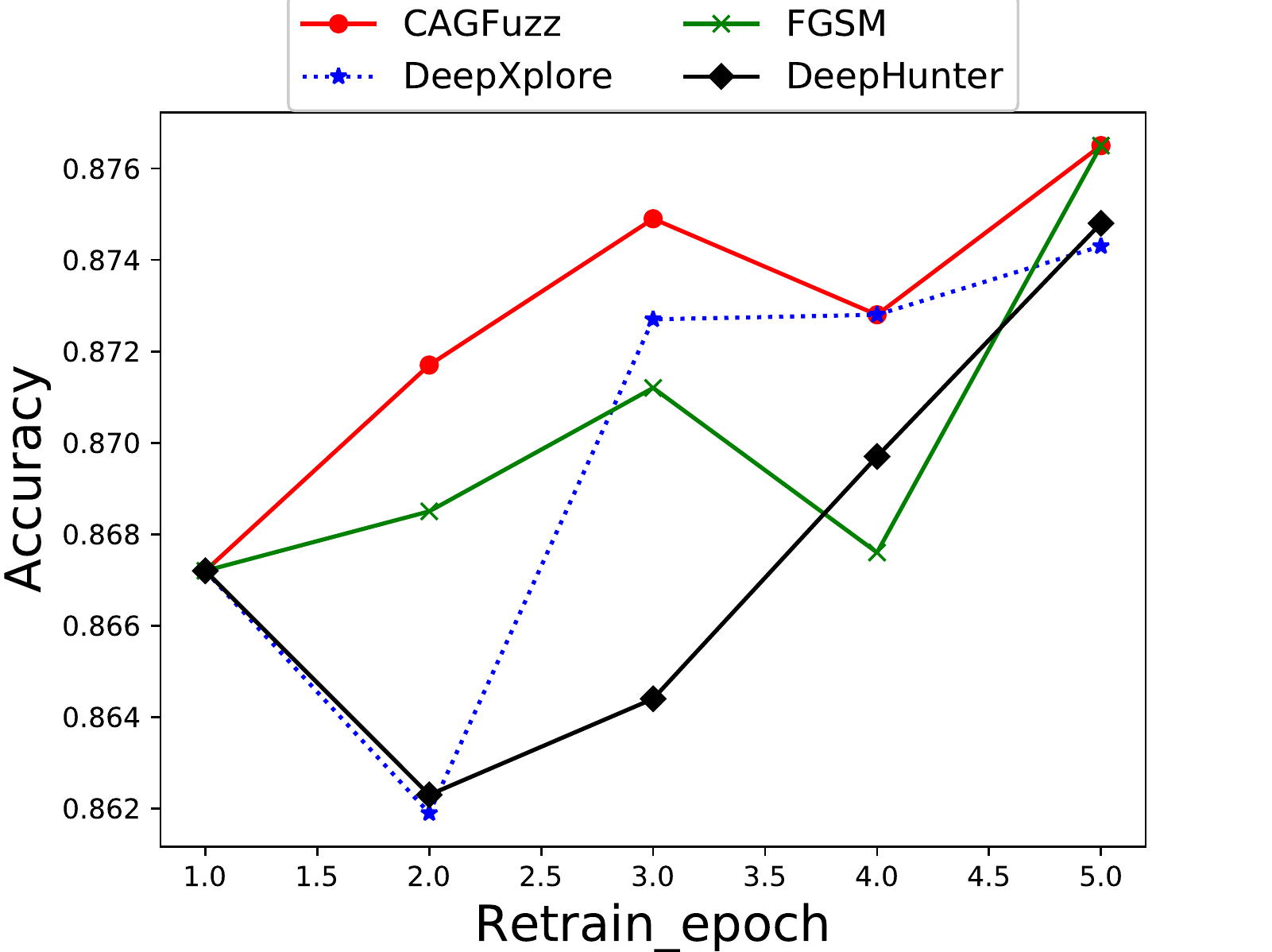}}
   \subfigure[VGG-19]{
   \label{figuree}
   \includegraphics[width=0.32\linewidth]{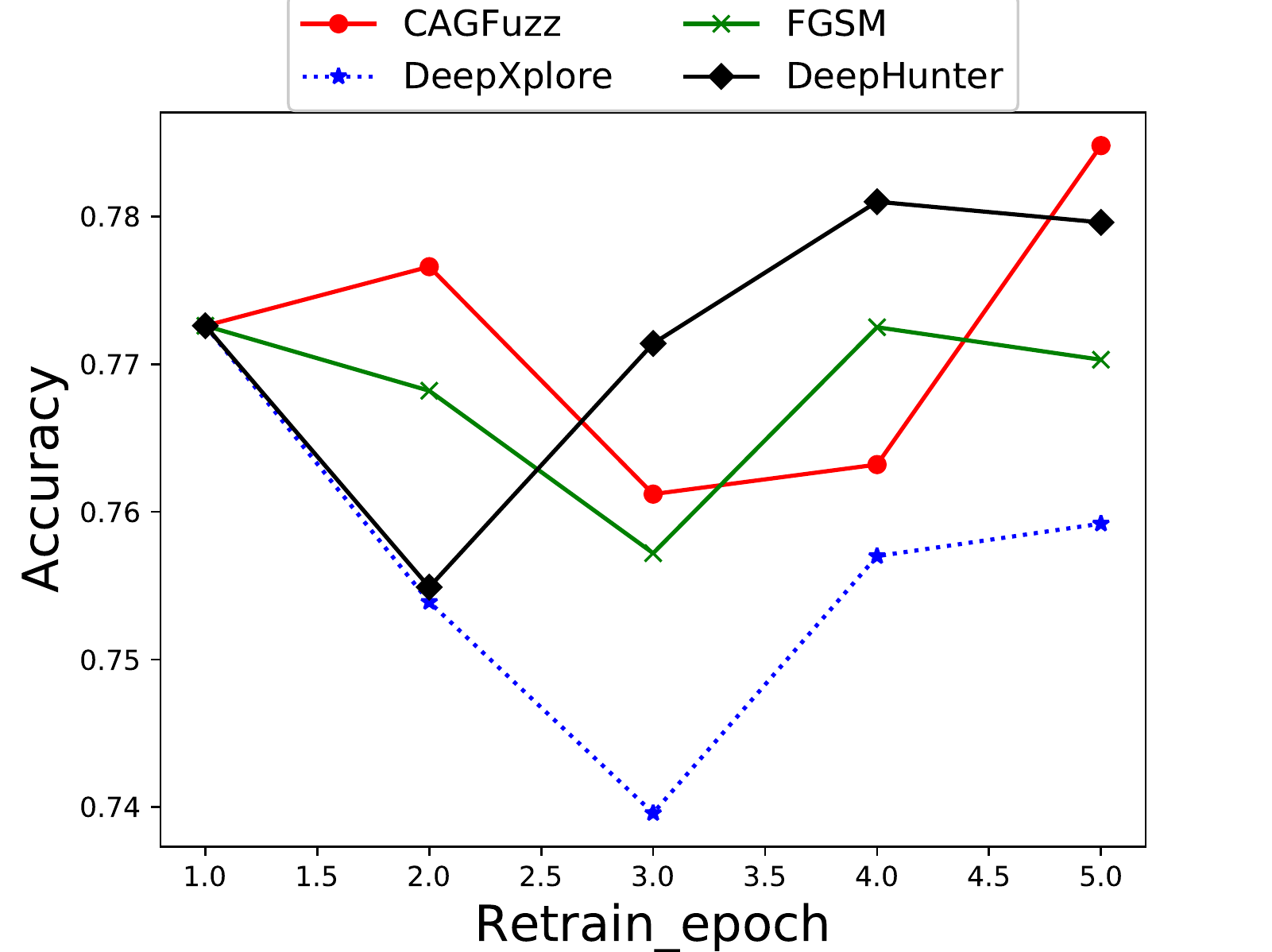}}
   \subfigure[ResNet-20]{
   \label{figuref}
   \includegraphics[width=0.32\linewidth]{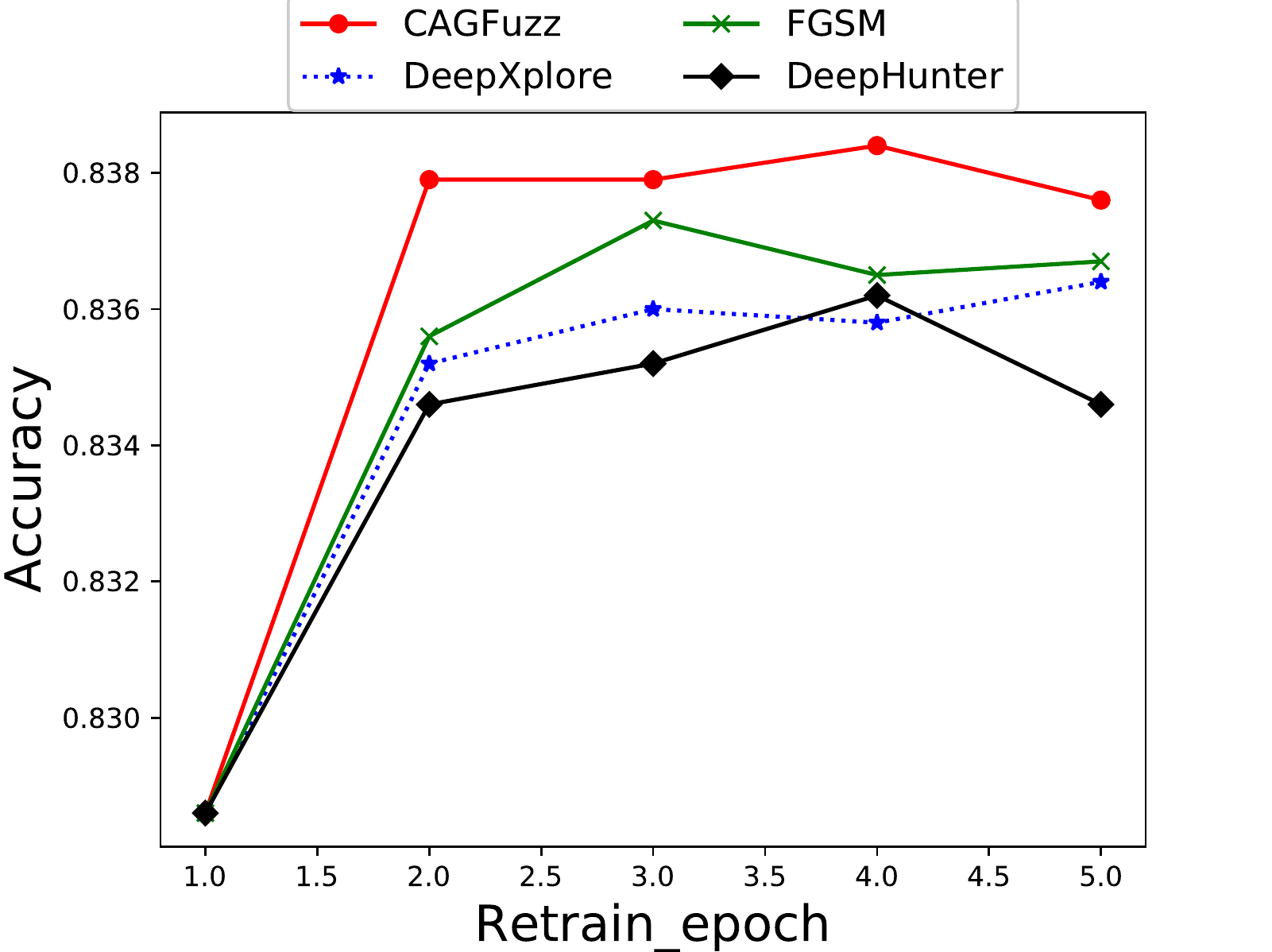}}
 \caption{Improvement in accuracy of DNN models when the training set is augmented with the same number of inputs generated by \emph{FGSM}, \emph{DeepXplore}, \emph{DeepHunter} and \emph{CAGFuzz}}.\label{Fig_DX}
\end{figure*}

To further validate our approach, we pre-train models on the MNIST and CIFAR-10 data sets. We further expand the training data by adding the same number of new generated examples, and train DNNs by 5 epochs. Our experiment results shown in Fig.~\ref{Fig_DX} are compared with other approaches. It can be found that \emph{CAGFuzz} sometimes has a low initial accuracy when the model is retrained. With the increase of epochs, the accuracy of the model increases rapidly, and the final accuracy for \emph{CAGFuzz} is higher than that of other approaches.

\emph{$\rhd$ Answer to \textbf{RQ4}:}  The accuracy of a DNN can be improved by retraining the DNN with adversarial examples generated by \emph{CAGFuzz}. The accuracy of the best model is improved from the original 86.72\% to 97.25\%, with an improvement of 12.14\%.

\subsection{Threats to Validity} \label{subsec_Threats}
In the following, we describe the main threats to validity of our approach in detail.

\textbf{Internal validity:} During the experimental process, the data set used to train \emph{AEG} is manually divided into two data domains, which may lead to subjective differences. To mitigate this threat, after the data domain is divided, we asked three observers to randomly exchange the examples of the two data domains, and three selected observers complete independently.
In addition, we pre-train with the initial data domains and then retrain with the data domains adjusted by other observers.

\textbf{External validity:}  During the experimental process, the classification of experimental data set is limited, which may lead to the reduction of the generality of the approach to a certain extent. To solve this problem, we use a cross-data set approach to validate the generalization performance of the approach.

\textbf{Conclusion validity:} According to the designed three problems, we can validate our approach. To further ensure the validity of the conclusion, we validated the conclusion through the valid data sets and models from other researchers, and reached the same conclusion as the standard data set.

\section{Related work}\label{sec_Related work}
In this section, we review the most relevant work in three aspects. Section~\ref{Adversarial_DL} introduces the adversarial deep learning and some adversarial examples generation approaches. Section~\ref{trad fuzztest} elaborates coverage-guided fuzz testing of traditional software. Section~\ref{testofDL} introduces the state-of-the-art testing approaches of DL systems.

\subsection{Adversarial Deep Learning}\label{Adversarial_DL}
A large number of recent research has shown that adversarial examples with small perturbations poses a great threat to the security and robustness of DL systems~\cite{Nguyen2015Deep,zhao2017generating,Samangouei2018Defense,yuan2019adversarial}. Small perturbations to the input images can fool the whole DL systems, and the input image is initially classified correctly by the DL systems. Although in human eyes, the modified adversarial example is obviously indistinguishable from the original example.

Goodfellow et al.~\cite{goodfellow2014explaining} proposed \emph{FGSM} (Fast Gradient Sign Method) which can craft adversarial examples using loss function $J(\theta ,x,y)$ with respect to the input feature vector, where $\theta$ denotes the model parameters, $x$ is the input, and $y$ is the output label of $x$. The adversarial example is generated as: $x^{'}=x+\epsilon sign(\triangledown _{x}J(\theta ,x,y))$. In this paper, we choose \emph{FGSM} as a baseline. The \emph{FGSM} approach uses the gradient change of specific DNN to generate adversarial examples. Consequently, the generated adversarial examples have good defect detection ability for the specific DNN. However, the approach cannot achieve good performance when it is extended to other DNNs.

Papernot et al.~\cite{papernot2016limitations} proposed JSMA (Jacobian-based Saliency Map Attack) to craft adversarial examples based on a precise understanding of the mapping between inputs and outputs of DNNs. For an input $x$ and a neural network $N$, the output of class $j$ is denoted as $N_{j}(x)$. To achieve a target misclassification class $t$, $N_{t}(x)$ is increased while the probabilities $N_{j}(x)$ of all other classes $j \neq t$ decrease, until $t=\arg max_{j}N_{j}(x)$.

Kurakin et al.~\cite{kurakin2016adversarial} proposed BIM (Basic Iterative method). They apply it multiple times with small step size, and clip pixel values of intermediate results after each step to ensure that they are in an $\epsilon$-neighbourhood of the original image. The method applies adversarial noise $\eta$ many times iteratively with a small parameter $\epsilon$, rather than one $\eta$ with one $\epsilon$ at a time, which gives a recursive formula:
$x_{0}^{'}=x$
and
$x_{i}^{'}=clip_{x,\epsilon }(x_{i-1}^{'}+\epsilon sign(\triangledown _{x_{i-1}^{'}}J(\theta ,x_{i-1}^{'},y))$,
where $clip_{x,\epsilon }(.)$ denotes a clipping of the values of the adversarial example such that they are within an $\epsilon$-neighborhood of the original input $x$.

Carlini et al.~\cite{carlini2017towards} proposed CW (Carlini and Wagner Attacks), a new optimization-based attack technique which is arguably the most effective in terms of the adversarial success rates achieved with minimal perturbation. In principle, the CW attack is to approximate the solution to the following optimization problem: $\arg min_{x^{'}}\lambda L(x,x^{'})-J(\theta ,x^{'},y)$, where $L$ is a loss function to measure the distance between the prediction and the ground truth, and the constant $\lambda$ is to balance the two loss contributions.


At present, these approaches are not used for testing deep learning systems. We find that it is meaningful to apply them to the steps of example generation in deep learning test. However, all these approaches only attempt to find a specific kind of error behavior, that is, to force incorrect prediction by adding minimum noise to a given example. In this way, these approaches are designed for special DNNs, and the generated adversarial examples have low generalization ability. In contrast, our approach does not depend on a specific DNN, and uses the distribution of general data domains to learn from each other, so as to add small perturbations to the original examples.

\subsection{Coverage-Guided Fuzzing Testing}\label{trad fuzztest}
Coverage-guided fuzzing testing (CGF)~\cite{GanCollAFL} is a mature defect and vulnerability detection technology. A typical CGF usually performs the following loops: 1) selecting seeds from the seed pool; 2) mutating seeds for a certain number of times to generate new tests using bit/byte flip, block substitution, and crossover of two seed files; 3) running the target program for the newly generated input and recording the execution trajectory; 4) if the detection is made in example of collapse, the fault seeds are reported and the interesting seeds covered with new traces are stored in the seed pool.

Superion~\cite{wang2019superion} conceptually extends LangFuzz~\cite{holler2012fuzzing} with coverage-guided: the seeds of structural mutation that increase coverage are retained to further fuzzing. While Superion works well for highly structured inputs such as XML and JavaScript, AFLSMART's variation operators better support block based file formats such as image and audio files.

Zest~\cite{padhye2018zest} and libprotobuf mutator~\cite{serebryany2017structure} have been proposed to improve the mutation quality by providing structure aware mutation strategies. Zest compiles the syntax specification into a fuzzer driver stub for the coverage-guided greybox fuzzer. This fuzzer driver translates byte-level mutations of LibFuzzer~\cite{serebryany2015libfuzzer} into structural mutations of the fuzzer target.

NEZHA~\cite{petsios2017nezha} is used to focus on inputs that are more likely to trigger logic errors by using behavioral asymmetries between test programs. The behavior consistency between different implementations acts as Oracle to detect functional defects.

\emph{TensorFuzz}~\cite{odena2018tensorfuzz} is good at automatically discovering errors that only a few examples can cause. For example, it can find the numerical error in the trained neural network, generate the difference between the neural network and its quantized version, and find the bad behavior in the character level language model. However, the defects of \emph{TensorFuzz} are as follows. First, \emph{TensorFuzz} directly adds noise to the examples, so it is unnatural to generate examples, while \emph{CAGFuzz} uses \emph{AEG} to mutate the examples first and then to restore them; thus, in \emph{CAGFuzz} the generated adversarial examples are more natural and understandable for humans. Second, \emph{TensorFuzz}  does not consider the deep feature while \emph{CAGFuzz} uses deep feature to constrain the adversarial examples; this enables us to ensure that the high-level semantics of the examples remain unchanged.

The validity of \emph{DLFuzz}~\cite{guo2018dlfuzz} shows that it is feasible to apply the fuzzy knowledge to DL testing, which can greatly improve the performance of existing DL testing technologies such as \emph{DeepXplore}~\cite{Pei2017DeepXplore}. Gradient-based optimization problem solution ensures simple deployment and high efficiency of the framework. The mechanism of seed maintenance provides different directions and more space for improving the coverage of neurons.

Due to the inherent difference between DL systems and traditional software, traditional CGF cannot be directly applied to DL systems. In our approach, CGF is adopted to be suitable for DL systems. The state-of-the-art CGF mainly consists of three parts: mutation, feedback guidance, and fuzzing strategy, in which we replace mutation with the adversarial example generator trained by \emph{CycleGAN}. In the feedback part, neuron coverage is used as the guideline. In the fuzzy strategy part, because the test is basically input by the same format of images, the adversarial examples generated with higher coverage are selected and put into the processing pool to maximize the neuron coverage of the target DL systems.

\subsection{Testing of DL Systems}\label{testofDL}
In traditional software testing, the main idea of evaluating machine learning systems and deep learning systems is to randomly extract test examples from manually labeled datasets~\cite{Witten2005Data} and hoc simulations~\cite{Rosenband2017Inside} to measure their accuracy. In some special cases, such as autopilot, special non-guided simulations are used. However, without understanding the internal mechanism of models, such black-box test paradigms cannot find different situations that may lead to unexpected behavior~\cite{Pei2017DeepXplore,goodfellow2017challenge}.

\emph{DeepXplore}~\cite{Pei2017DeepXplore} proposes a white-box differential testing technique for generating test inputs that may trigger inconsistencies between different DNNs, which may identify incorrect behavior. For the first time, this method introduced concept of neuron coverage as a metric of DL testing. At the same time, it requires multiple DL systems with functions similar to cross-reference prediction to avoid manual checking. However, cross-references have difficulties in finding DL-like systems. \emph{DeepXplore} is similar to \emph{FGSM} in that it is also based on the given DNN model to generate adversarial examples, whose generalization ability is not good. Furthermore, through our experiments, we found that \emph{DeepXplore} has a poor ability on finding potential errors in the model. The reason may be that \emph{DeepXplore} does not use any constraints to control the generation of adversarial examples.
In contrast, our approach, \emph{CAGFuzz} uses \emph{AEG} to generate adversarial examples based on a given data set. Experiments show that the generalization ability of these adversarial examples is better. In addition, we use deep feature to constrain the generation of adversarial examples, which has a good effect in finding potential errors of the model.

\emph{DeepHunter}~\cite{Xie2018DeepHunter} performs mutations to generate new semantic retention tests, and uses multiple pluggable coverage criteria as feedback to guide test generation from different perspectives. Similar to traditional coverage-guided fuzzy (CGF) testing~\cite{Pham2016Coverage,Wang2017Skyfire}, \emph{DeepHunter} uses random mutations to generate new test examples. Although there is a screening mechanism to filter invalid use examples, it still wastes time and computing resources. \emph{DeepHunter} uses pixel value transformation (change image contrast, image brightness, image blur and image noise) and affine transformation (image translation, image scaling, image shearing, and image rotation) to mutate the image. The examples generated by these image transformations are unnatural, and the human eye can clearly see the components of ``fraud". In addition, \emph{DeepHunter} uses pixel level constraints when keep valid examples, which are the low-level features of the image. When testing the model with deeper layers, the test effect is not good. In contrast, in \emph{CAGFuzz}, \emph{AEG} generates adversarial examples by adding small perturbations to the original examples that are not visible to the human eye, and the adversarial examples generated by \emph{AEG} are natural and more confusing to the model. At the same time, the deep features are used in \emph{CAGFuzz} to constrain the adversarial examples, and consequently the adversarial examples have effective test effect on the model with deep layers.

\emph{DeepTest}~\cite{tian2018deeptest} performs a tool for automated
testing of DNN-driven autonomous cars. \emph{DeepTest} does not consider the small perturbations on input examples and however maximizes the neuron coverage of a DNN using synthetic test images generated by applying different real transformations to a set of seed images. The image transformation method of \emph{DeepTest} is the affine transformation (image translation, image scaling, image shearing and image rotation) in \emph{DeepHunter}. Therefore, \emph{DeepTest} has the similar problem as \emph{DeepHunter}.

In addition, many testing approaches for traditional software have also been adopted and applied to testing DL systems, such as
MC/DC coverage~\cite{Sun2018Testing}, concolic test~\cite{sun2018concolic}, combinatorial test~\cite{ma2018combinatorial} and mutation test~\cite{ma2018deepmutation}. Furthermore, various forms of neuron coverage~\cite{Lei2018DeepGauge} have been defined, and are demonstrated as important metrics to guide test generation. In general, these approaches do not consider adversarial examples and test DL systems from other aspects.

\section{Conclusions and suggestions for future work}\label{sec_conclusions}
 We design and implement \emph{CAGFuzz}, a coverage-guided adversarial generative fuzzing testing approach. \emph{CAGFuzz} trains an adversarial example generator for a specified dataset. It generates adversarial examples for target DNN by iteratively taking original examples, generating adversarial examples and feedback of coverage rate, and finds potential defects in the development and deployment phase of DNN. We have done a lot of experiments to prove the effectiveness of \emph{CAGFuzz} in promoting DNN coverage, discovering potential errors in DNN and improving model accuracy. The goal of \emph{CAGFuzz} is to maximize the neuron coverage and the number of potential erroneous behaviors. The experimental results show that \emph{CAGFuzz} can detect thousands of erroneous behaviors in advanced DNN models, which are trained on publicly popular datasets.

Several directions for future work are possible.
 \begin{itemize}
\item At present, we only use neuron coverage to guide the generation of adversarial examples. Neuron coverage may not cover all the logic of DNN effectively. In the future, we can use multidimensional coverage feedback to improve the information that adversarial examples can cover.
  \item \emph{CAGFuzz} adds perturbation information to the original example by mapping between two data domains. These perturbations are uncontrollable. In the future, the perturbation information can be added to the original example by feature control.
 \item This paper mainly studies image examples, and how to train effective adversarial example generator for other input forms, such as text information and voice information, is also a meaningful direction.
 \end{itemize}

\section{Acknowledgements}
The work is supported by the National Natural Science Foundation
of China under Grant No. 61572171, the Natural Science Foundation of Jiangsu Province under Grant No. BK20191297, and the Fundamental
Research Funds for the Central Universities under Grant No. 2019B15414.

\bibliographystyle{ieeetr}
\bibliography{reference}

\begin{IEEEbiography}[{\includegraphics[width=1in,height=1.25in,clip,keepaspectratio]{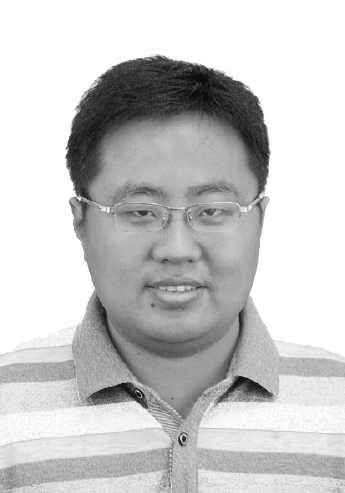}}]{Pengcheng Zhang}
received the Ph.D. degree in computer science from Southeast University in 2010. He is currently an associate professor in College of Computer and Information, Hohai University, Nanjing, China, and was a visiting scholar at San Jose State University, USA. His research interests include software engineering, service computing and data mining. He has published in premiere or famous computer science journals, such as IEEE TBD, IEEE TETC, IEEE TSC, IST, JSS, and SPE. He was the co-chair of IEEE AI Testing 2019 conference. He served as technical program committee member on various international conferences. He is a memeber of the IEEE.
\end{IEEEbiography}

\begin{IEEEbiography}[{\includegraphics[width=1in,height=1.25in,clip,keepaspectratio]{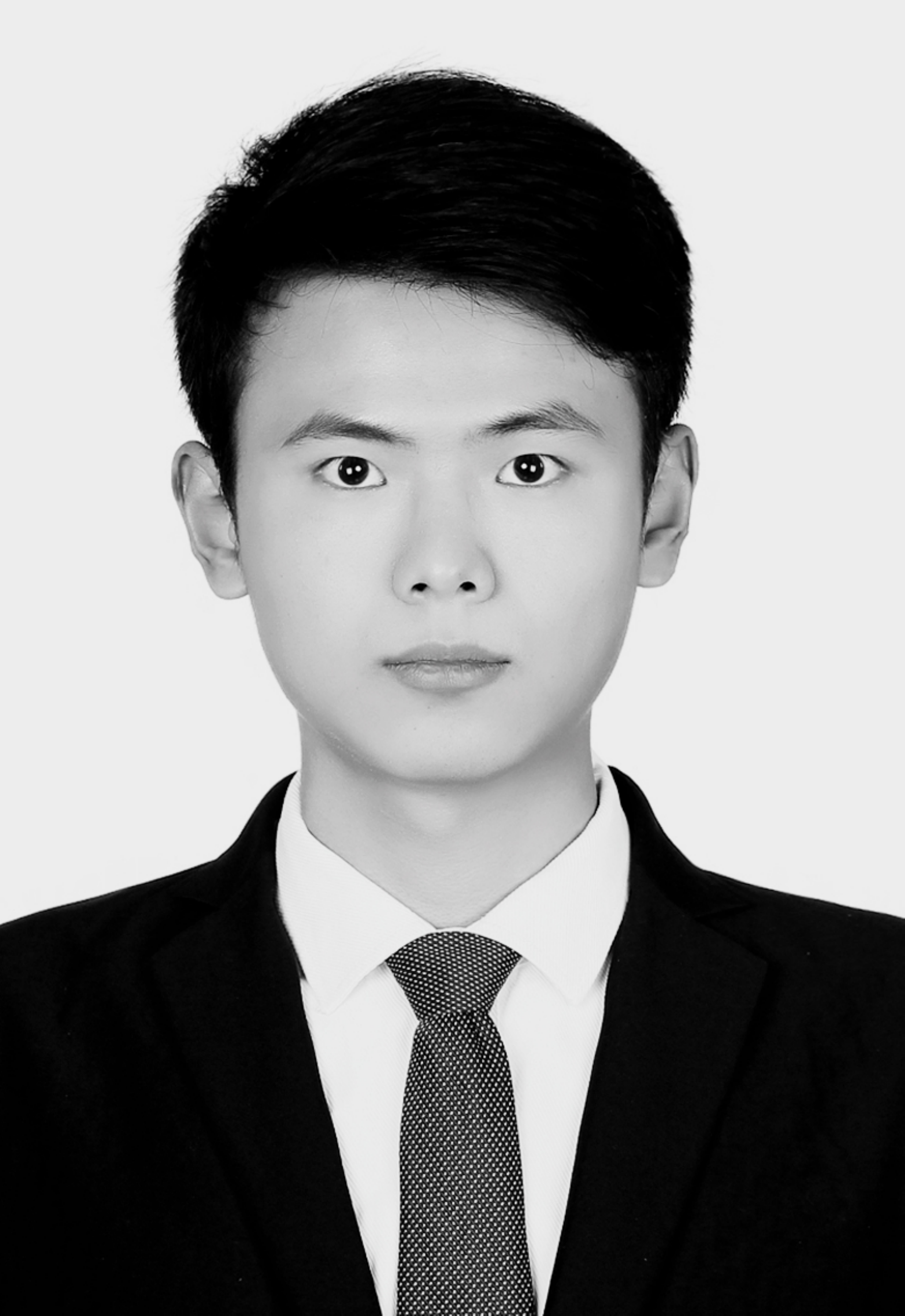}}]{Qiyin Dai} received the bachelor's degree in computer science and technology from nanjing university of finance and economics in 2018. He is currently working toward the M.S. degree with the College of Computer and Information, Hohai University, Nanjing, China. His current research interests include data mining and software engineering.
\end{IEEEbiography}
\begin{IEEEbiography}[{\includegraphics[width=1in,height=1.25in,clip,keepaspectratio]{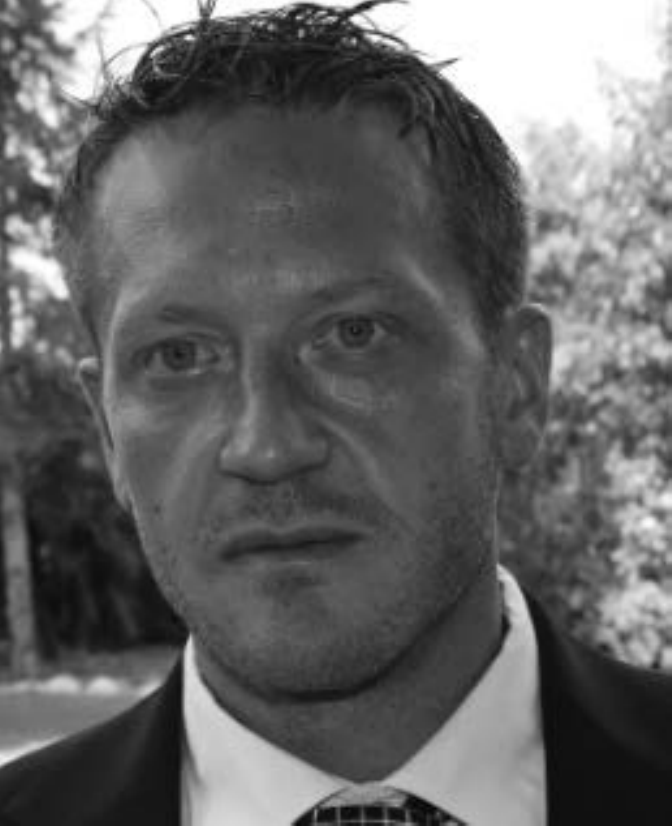}}]{Patrizio Pelliccione}
is Associate Professor at the Department of Information Engineering, Computer Science and Mathematics - University of L'Aquila (Italy), and he is also Associate Professor at the Department of Computer Science and Engineering at Chalmers University of Gothenburg (Sweden). He got his PhD in 2005 at the University of L'Aquila (Italy) and from February 1, 2014 he is Docent in Software Engineering, title given by the University of Gothenburg (Sweden). His research topics are mainly in software engineering, software architectures modelling and verification, autonomous systems, and formal methods. He has co-authored more than 130 publications in journals and international conferences and workshops in these topics. He has been on the program committees for several top conferences, he is a reviewer for top journals in the software engineering domain, and he organized as program chair international conferences. He is very active in European and National projects. More information is available at \url{http://www.patriziopelliccione.com}.
\end{IEEEbiography}\textbf{}

\end{document}